# Generative AI in the Construction Industry: Opportunities & Challenges

**Prashnna Ghimire [1*], Kyungki Kim [1], Manoj Acharya[2]**

[1] Ph.D. Student, Durham School of Architectural Engineering & Construction, University of Nebraska-Lincoln, USA; pghimire3@huskers.unl.edu
[1] Assistant Professor, Durham School of Architectural Engineering & Construction, University of Nebraska-Lincoln, USA; kkim13@unl.edu
[2] AI Scientist, SRI International, USA; manoj.acharya@sri.com
* Correspondence: pghimire3@huskers.unl.edu;

**Abstract:** In the last decade, despite rapid advancements in artificial intelligence (AI) transforming many industry practices, construction largely lags in adoption. Recently, the emergence and rapid adoption of advanced large language models (LLM) like OpenAI's GPT, Google's PaLM, and Meta's Llama have shown great potential and sparked considerable global interest. However, the current surge lacks a study investigating the opportunities and challenges of implementing Generative AI (GenAI) in the construction sector, creating a critical knowledge gap for researchers and practitioners. This underlines the necessity to explore the prospects and complexities of GenAI integration. Bridging this gap is fundamental to optimizing GenAI's early-stage adoption within the construction sector. Given GenAI's unprecedented capabilities to generate human-like content based on learning from existing content, we reflect on two guiding questions: What will the future bring for GenAI in the construction industry? What are the potential opportunities and challenges in implementing GenAI in the construction industry? This study delves into reflected perception in literature, analyzes the industry perception using programming-based word cloud and frequency analysis, and integrates authors' opinions to answer these questions. This paper recommends a conceptual GenAI implementation framework, provides practical recommendations, summarizes future research questions, and builds foundational literature to foster subsequent research expansion in GenAI within the construction and its allied architecture & engineering domains.

## 1. Introduction

In the last four decades, the field of machine learning (ML), particularly the deep learning subdomain reliant on artificial neural networks, has undergone substantial maturation, causing immense transformations across many industrial landscapes [1]. It has emerged as a powerful asset, automating procedures within the construction sector, an industry that trails behind others in both efficiency and output. However, embracing this paradigm shift faces impediments due to gradual headway in overseeing data quality and the absence of directives for integrating domain expertise with data-centric evaluation. These challenges crystallize into three critical concerns: the disparity between a feature-rich space and limited samples, the balance between model precision and applicability, and the reconciliation of machine learning outcomes with field-specific insights [1], [2]. Here are three simple examples of these challenges: (1) A construction company has a large amount of data on the features of construction projects, but only data on a limited number of projects. This disparity between the feature-rich space and the limited samples makes it difficult to train a machine learning model that can precisely predict the cost of construction projects, (2) An owner organization is trying to implement a machine learning model to predict the completion time of a construction project based on data they have access to such as project value, delivery method, complexity, and materials quantity in previous projects. However, the company wants to make sure that the model is applicable to a wide range of projects, so it does not want to make the model too precise. A more precise model will be able to make more accurate predictions about the completion time of a project, but it may not be applicable to a wide range of projects. A less precise model will be more applicable to a wider range of projects, but it may not be as accurate, (3) Safety manager is using a machine learning model to predict the likelihood of a fall accident on a construction site and has access to data on the weather, the type of construction, and the safety practices used on previous projects and predicts that there is a 10% chance of a fall accident on the current project. However, the developed model may not be able to account for all of the factors, such as human errors, unforeseen conditions, that can contribute to an accident. Therefore, traditional machine learning algorithms are somewhat constrained in their capabilities restricted to these limitations[3].

The rapid growth of artificial intelligence (AI), a discipline that involves developing computer systems capable of human-like cognition and actions, has enabled the advancement of sophisticated large language models (LLMs), such as GPT, PaLM, and Llama. GenAI, a subset of deep learning, leverages neural networks, and can process both labeled and unlabeled data using supervised, unsupervised, and semi-supervised methods to synthesize novel content like text, images, and audio [4], [5]. An LLM trains models on existing data, constructing statistical representations to predict content. When provided prompts, generative systems output new synthesized content learned from underlying patterns. Architecturally, transformer models enable GenAI, containing encoders to process inputs and decoders to translate them into contextually relevant outputs [5]. There are four major types of GenAI models: text-to-text, text-to-image, text-to-video/3D, and text-to-task. Text-to-text models, trained to learn mappings between text pairs, accept natural language input and generate text output [6]. Text-to-image models, a recent development, are trained on image datasets paired with text captions. These models take text prompts as input and generate corresponding images as output, often using diffusion techniques[7]. Text-to-video models synthesize videos from text prompts, accepting inputs ranging from single sentences to full scripts, and outputting corresponding video representations [8]. Similarly, text-to-3D models create 3D objects that match a user's textual description. Text-to-task models are trained to execute particular tasks based on textual prompts. These models can perform diverse actions including responding to questions, conducting searches, making predictions, and carrying out requested behaviors [9]. LLMs are a type of general AI. As large pre-trained models designed for adaptability, foundation models like GPT constitute AI architectures that are trained on vast data quantities. This enables fine-tuning to a wide range of tasks including question answering (Q&A), sentiment analysis, information extraction, image captioning, object recognition, instruction following, and more. [10]

Over the past few decades, in the construction, researchers have published articles on implementing AI and its subdomains to address industry-specific challenges. These studies demonstrate AI and machine learning applications across the construction management spectrum, including safety management [11]–[15], cost predictions [16]–[20], schedule optimization [1], [21], [22], progress monitoring [23]–[27], quality control [28], [29], supply chain management[30]–[33], logistics management[34], [35], project risks management [36]–[41], disputes resolution [42], [43], waste management [44]–[46], sustainability assessments[47]–[51], visualization [52], [53], and overall construction process improvements [1], [54]–[57]. Also, there have been studies highlighting the integration of AI with Building Information Modeling (BIM) to enhance information extraction, streamline workflows, and optimize construction management efficiency [58]–[62]. Furthermore, some research studies also emphasized the impact of robotics and AI integration in construction such as improvements in construction quality, safety, project acceleration, and the mitigation of labor shortages [63]–[66]. However, there is a noticeable gap in research on GenAI's applications, future opportunities, and adoption barriers specific to the construction industry. This gap is likely due to the recent and rapid emergence of GenAI as a novel technology for this field, resulting in a delay in research and implementation when compared to other industries that have already begun to explore and capitalize on the benefits of GenAI adoption [2], [4], [67]–[69], [70]. As the construction industry continues to deal with its unique challenges, there exists a vital need to bridge this research gap, uncover the untapped opportunities offered by GenAI, and address the barriers obstructing its adoption within the construction sector.

With this background, in this study we seek to answer the two major research questions: (1) What are the current opinions and evidence about the opportunities & potential applications, and overall challenges related to GenAI technologies implementation in the context of construction?, and (2) What are the most important research questions to investigate in future related to GenAI technologies in the context of construction? The remainder of this paper is arranged as follows: Section 2 summarizes our methodology. Section 3 describes various GenAI model structures and presents related work in construction. Section 4 synthesizes opinions and evidence on opportunities, summarizes potential application areas, and visualizes conceptual implementation framework, and Section 5 examines key challenges, from technical limitations to industry challenges. The recommendations for implementation, and critical research questions to prioritize investigating GenAI's unknowns in construction will be discussed in Section 6. Finally, Section 7 concludes by spotlighting this study's significant findings.

## 2. Methodology

To achieve our research goals, we followed a research framework as mentioned in Figure 1. Given the limited literature on generative AI in construction, we conducted a non-systematic review using keywords like "Generative AI AND Construction", "Generative AI", and "Large Language Models AND Construction" in Scopus and Google Scholar. We then used the snowball method, identifying key articles and mining their references and citations to find more relevant studies. In addition, to get the most up-to-date insights, construction industry professionals' perceptions of generative AI via posts on LinkedIn over the three months leading up to August 20, 2023. Using three keyword combinations - "Generative AI in construction", "#generativai #construction", and "#generativeai #aec" - we identified 32 relevant opinions comprising a total of 63,778 words. Our analysis incorporated various formats including posts, comments, polls, and articles. Articles accounted for 48% of the data, comments 34%, posts 16%, and polls 6%. To analyze this data, we utilized programming-based text mining techniques including word cloud analysis to highlight the most frequent terms, sentiment analysis to categorize opinions as positive, negative, or neutral, and frequency analysis to summarize key themes throughout the corpus. With a literature review and industry perspectives, this paper outlines potential GenAI applications in construction. A conceptual implementation framework is then proposed to implement identified applications, along with key implementation challenges.

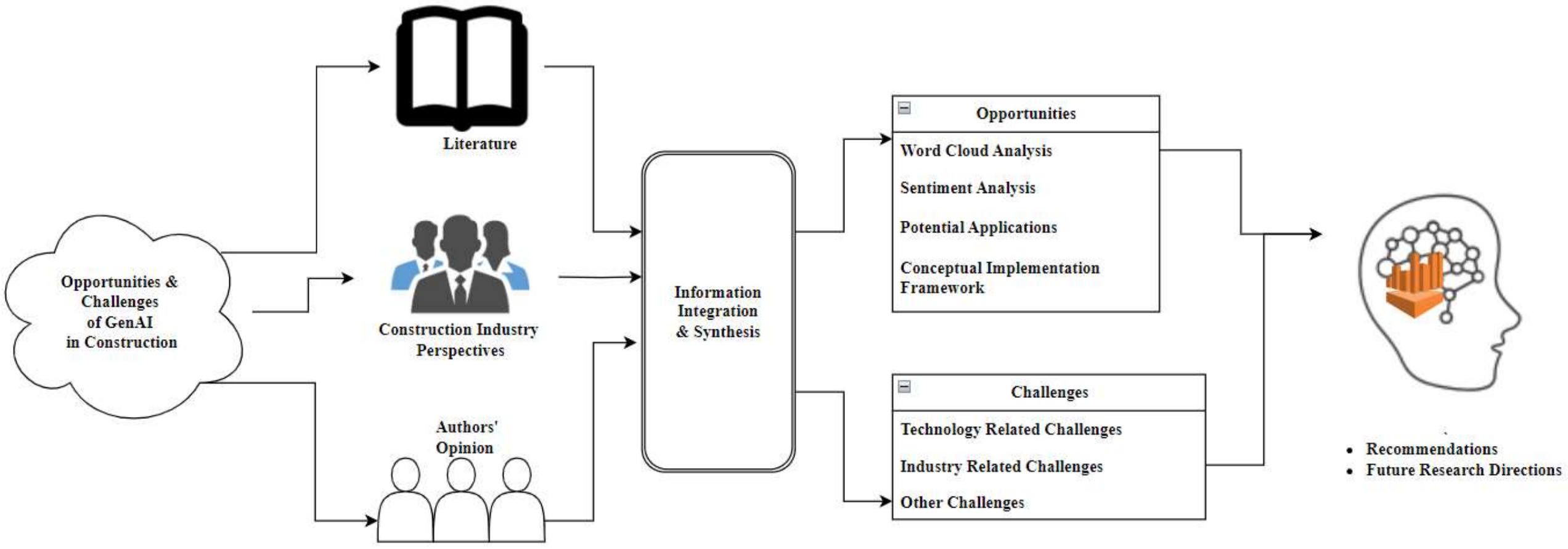

Figure 1. Research Framework

Furthermore, we integrated the perspectives of the authors in this study. As experts in allied disciplines related to emergence technology such as generative AI in the built environment, the authors contribute more than a decade of combined experience in areas including AI in construction, automation in construction, and generative AI specifically.

## 3. Various GenAI Model Structures and Related Work in Construction

In recent years, researchers have increasingly focused on modifying the learning algorithms of generative AI (GenAI) models to fit specific domains and tackle industry-specific problems. The choice of which generative AI model to use depends on the specific task at hand. Based on their generative mechanism, there are five major types of GenAI models [2], [71], [72]. Generative Adversarial Networks (GAN) are often used for image generation because they can create realistic images. Variational AutoEncoders (VAE) are commonly used for text generation, as they can produce clear, grammatically correct samples by learning the original distribution of the training data. Autoregressive models are best at text generation similar to their training data, since they generate text token-by-token while conditioning on previous tokens. Diffusion models can create smooth and natural image samples by starting with noise and reversing a diffusion process. And, flow-based models learn transformations between data and latent representations, enabling diverse and creative image generation. In the following subsections, we will investigate the background of each model, explain their operational mechanisms including model architecture, underline any limitations, examine their relevance within the construction domain, if such use cases exist, and summarize the characteristics, advantages, and disadvantages of all models.

### 3.1. Generative Adversarial Network

First introduced by Goodfellow et al. in 2014, GANs are a type of deep learning model comprised of two neural networks: a generator and a discriminator[73]. The generator is tasked with creating new synthetic data, while the discriminator attempts to differentiate between real and generated data. As shown in Figure 2 (a) [72], GANs are trained through an adversarial process, where the generator produces fake samples that are fed along with real samples into the discriminator. The discriminator then predicts which samples are real or fake, and loss gradients are calculated using a loss function to update both models. During training, the generator tries to fool the discriminator by improving its ability to generate realistic data [71], [74]. The format of the real and synthetic data samples can vary, as long as the neural network architectures are adapted accordingly. GANs have proven adept at generating images, video, and text that are remarkably close to actual data distributions. Their adversarial training process allows for modeling complex, multi-modal data. However, GAN training can be unstable, and finding the optimal balance between the generator and discriminator is challenging [75].

GANs have shown possibilities for a variety of applications in the construction industry. Researchers have demonstrated that GANs can generate plausible technical drawings, including floorplans, mechanical/electrical/plumbing diagrams, sectional views, and colored plans [72]. The adversarial training process allows GAN models to synthesize images that closely match the style and content of real architectural drawings across multiple domains. In another study, GANs have been applied to generate photorealistic renderings of building facades [76]. By learning from datasets of real facade images, GANs can produce synthetic views that are useful for tasks like style classification and image restoration.

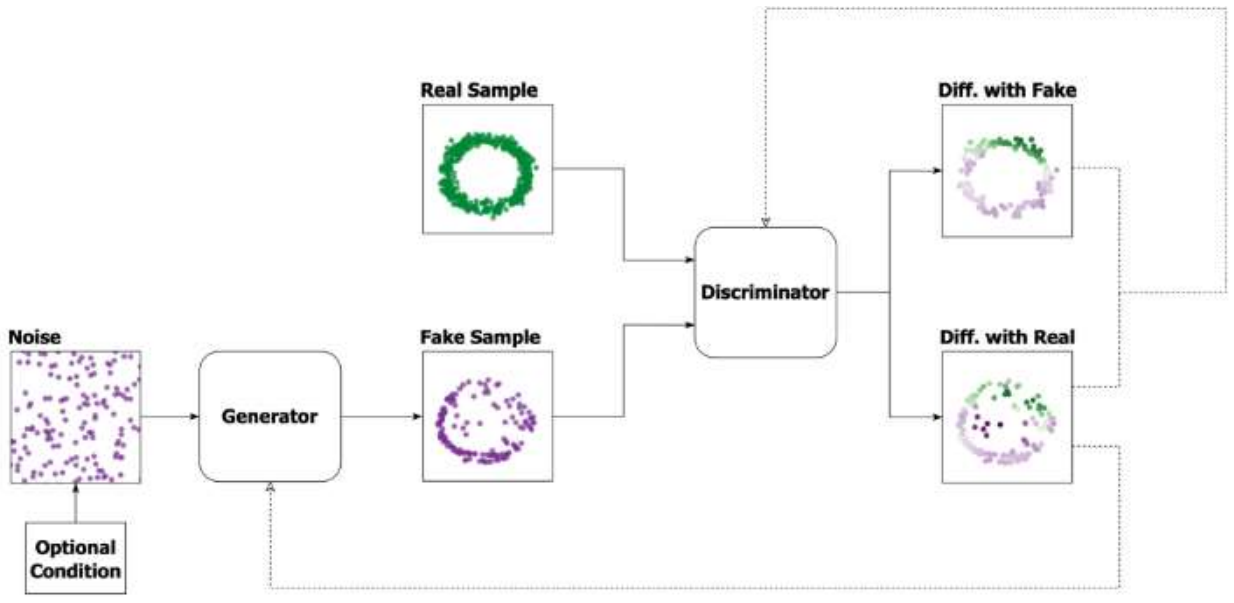

Figure 2 (a). General GAN [72]

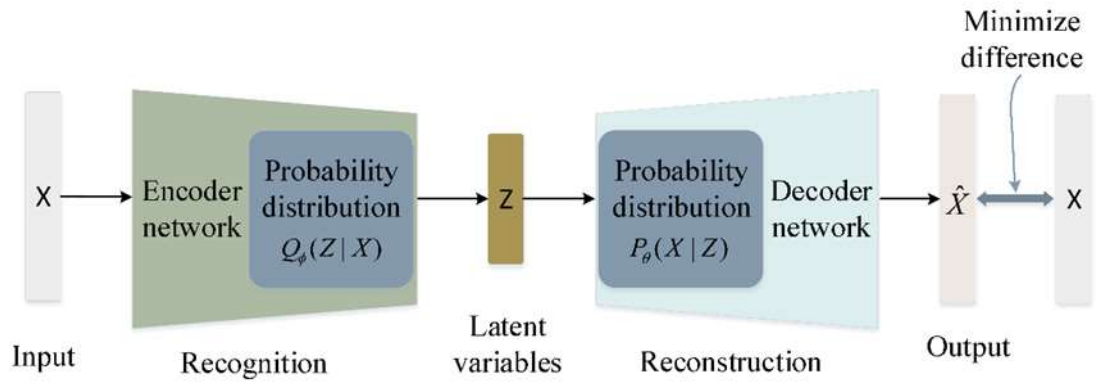

Figure 2 (b). VAE Model[77]

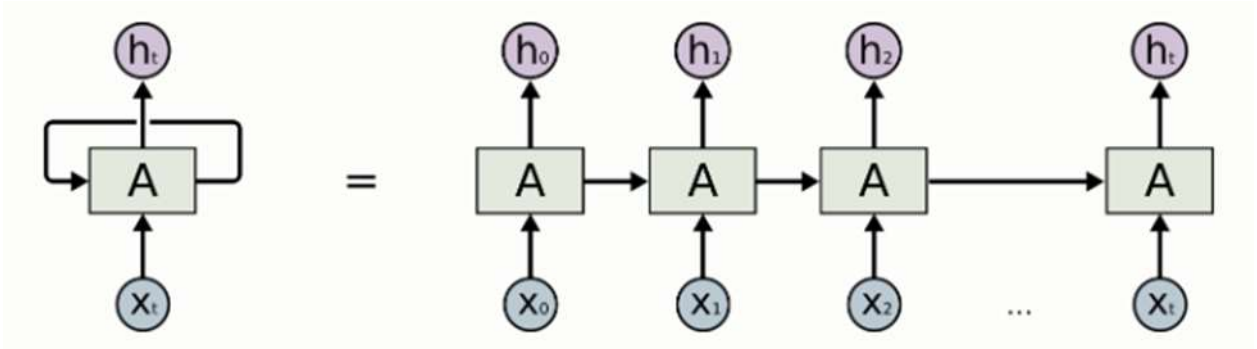

Figure 2 (c). RNN Diagram [78]

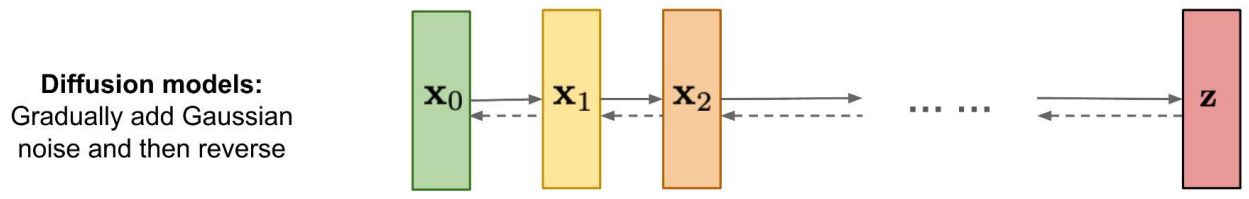

Figure 2 (d). Diffusion Model [79]

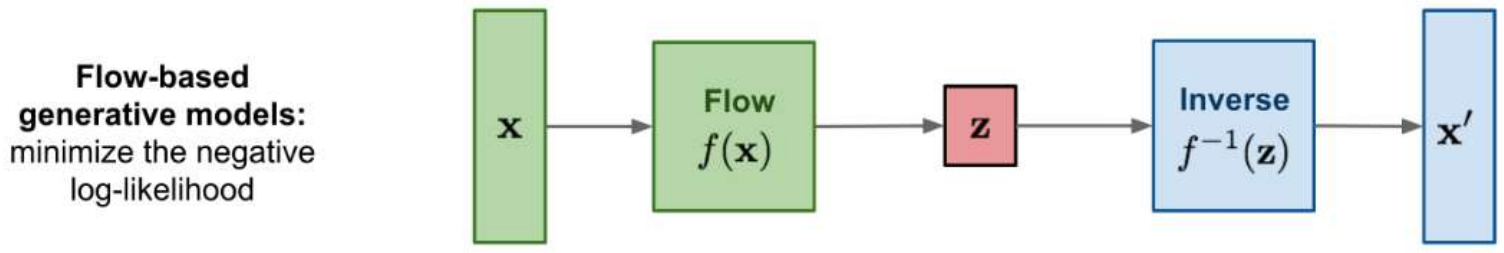

Figure 2 (e). Flow-based Model[80]

Figure 2. GenAI Models

### 3.2. Variational AutoEncoders

Variational Autoencoders (VAEs) are a class of generative models specifically designed to acquire a data representation in a lower-dimensional latent space. This latent space provides a compressed yet essential feature representation of the original data [81]. Kingma and Welling introduced VAEs in 2013, establishing them as a pivotal model in the field [82]. VAEs consist of two intertwined and independently parameterized components: the encoder, responsible for recognition, and the decoder, focused on generation. These components work in tandem to support each other's operations [83]. The model comprising an encoder network $Q_\phi(Z|X)$ and a decoder network $P_\theta(X|Z)$ is illustrated in Figure 2 (b). VAEs are proficient in approximate inference and can be effectively trained using gradient descent methods. The encoder network, characterized by parameters $\phi$, efficiently compresses data into the lower-dimensional latent space, mapping input data X to a continuous latent variable Z. Conversely, the decoder network, parameterized by $\theta$, utilizes this latent variable to generate data, performing the reverse mapping from Z to reconstructed data. Both the encoder and decoder employ deep neural networks for their construction, with parameters $\theta$ and $\phi$, respectively [77]. VAEs are trained to utilize variational inference, enabling the acquisition of a probabilistic distribution over the latent space. This learned distribution empowers VAEs to generate new data samples that closely resemble the training data. VAEs exhibit versatility and find applications in several domains, including data compression, image synthesis, text generation, and discovery. Because VAE imposes assumptions about the latent space, they are less flexible than other generative models in capturing complex real-world data distributions and data sequences [84], [85].

Like other industries, construction struggles with limited access to large datasets, a major obstacle for implementing deep learning models. While several studies have investigated big data challenges, solutions remain needed to compile requisite construction data. A recent study by Delgado & Oyedele [86] highlighted the approaches to addressing limited data including data augmentation through distortions and variants of original data, synthetic data generation with methods like VAE, and transfer learning. And, the study explored using VAE to expand financial datasets for construction projects, as financial data lacks the transformation invariance present in images, making AutoEncoders a promising technique. The results showed that VAE provided more robust outputs and better represented the non-linear correlations between the variables in the financial datasets. Another study by Balmer et al. [87] presented the use of VAEs for the conceptual design of pedestrian bridges from synthetically generated data, eliminating manual and time-consuming traditional design processes. Variational AutoEncoders show promise for generating new design and construction data to address limited datasets, and facilitating advanced deep learning applications. VAEs can be used to generate new data that is similar to existing data for defect detection, extract features from sensor data for predictive maintenance, model uncertainty in construction projects for risk assessment, and generate new designs for buildings or infrastructure. VAEs can learn from data at different levels of abstraction, depending on the specific task being performed.

### 3.3. Autoregressive models

An autoregressive model is a type of generative model that predicts the next token in a sequence, given the previous tokens. This means that the model is trained on a sequence of data, and it learns to predict the next token in the sequence based on the previous tokens[88]. One common architecture for an autoregressive model is a recurrent neural network (RNN) as shown in Figure 2 (c). The output at time ’t’ in an autoregressive model relies not only on the input '$x_t$' but also on prior inputs 'x' from preceding time steps. Nevertheless, in contrast to an RNN, the preceding 'x's are not conveyed through a concealed state; rather, they are directly supplied to the model as additional inputs [78]. Autoregressive generative models leverage the chain rule of probability to decompose the joint distribution of a sequence into conditional distributions over tokens based on their context [84], [89]. While autoregressive models are powerful density estimators, their sequential sampling is slow for high-dimensional data and requires a fixed ordering to decompose the data, which is not always straightforward [84].

A study by Elfahham [90] found the prediction of the construction cost index using the autoregressive time series method was most accurate compared to neural network and linear regression approaches. The autoregressive technique's specialized modeling of temporal dependencies allowed it to outperform. Autoregressive models have the potential to enable advanced analytics in construction by modeling temporal dependencies in historical data. Applications include forecasting construction costs, risk identification, schedule optimization, and automating tasks. These models capture relationships over time to predict future outcomes and empower data-driven decision-making.

### 3.4. Diffusion Models

Diffusion models, a type of GenAI, produce high-quality synthetic images and videos by learning to reverse an artificial diffusion process. This process involves gradually adding Gaussian noise to training data over multiple time steps, following a predefined schedule that gradually masks the original data[7], as shown in Figure 2 (d) [79]. During training, the model learns to take a noisy sample from an intermediate point within this noise schedule and subsequently predict a less noisy version of the data from the previous time step. By repeatedly applying this de-noising prediction across many time steps, the model can start from pure noise and reverse the diffusion back to a realistic generated image[91]. Though sampling is relatively slow due to the multiple required predictions, diffusion models can generate sharp and coherent outputs, especially for image generation. Their ability to condition the sampling makes them versatile and broadly applicable across computer vision tasks. Popular GenAI models like DALL-E2 and Imagen are based on the diffusion model concept[7]. Some studies underline the major limitations of the diffusion models such as poor time efficiency during inference requiring many evaluation steps, and high computational expense for the iterative de-noising [92], [93].

### 3.5. Flow-based Models

Flow-based models represent a category of GenAI models that generate synthetic outputs by framing the data generation process as a continuous normalizing flow. They work by taking noise vectors and repeatedly transforming them through a series of bijective functions, each designed to bring the distributions closer to the target data distribution. Unlike other generative models, the flow model only uses a reversible encoder to complete the model's construction, which makes the design more delicate [2] as shown in Figure 2 (e) [90]. Through these transformations, flow models can convert noise inputs into realistic generated samples. The origin of flow-based generative models dates back to the work of Dinh et al. in 2014 [94]. These models offer various advantages, including precise latent-variable inference, accurate log-likelihood evaluation, and efficiency in both inference and synthesis processes [95]. These models were further refined and extended by Dinh et al. in 2016 [96]. The flow-based models have some challenges in terms of training complexity due to the need for inverting networks and computing determinants, which creates a primary drawback.

Table 1 provides a summary of GenAI model types, their characteristics, advantages, and disadvantages. It helps in understanding and selecting the suitable generative model for specific applications.

Table 1. Summary of GenAI Models

| ***GenAI Model Type*** | **Characteristics** | **Advantages** | **Disadvantages** |
|---|---|---|---|
| *Generative Adversarial Network (GAN)* | Two neural networks, a generator, and a discriminator, compete with each other to generate realistic data. | - Generate high-quality data that is indistinguishable from real data. | - Unstable to train<br>- Difficult to find the right balance between the generator and discriminator. |
| *Variational AutoEncoder (VAE)* | Encodes data into a latent space and then decodes it back into the original space. | - Generate data that is similar to the training data. | - Less flexible than GANs<br>- Lack the ability to tackle sequential data<br>- Difficult to control the quality of the generated data. |
| *Autoregressive models* | Generate data one step at a time, using the previously generated data as input. | - Generate data that is very realistic, especially for text and speech. | - Slow to generate data in high dimension<br>- Difficult to scale to large datasets. |
| *Diffusion models* | Start with a noisy image and gradually refine it to a realistic image. | - Generate high-quality images from a small amount of data.<br>- Can be trained without paired or labeled datasets | - Slower generation process<br>- Computationally expensive |
| *Flow-based models* | Transform data from one distribution to another using a series of invertible functions. | - Flexible and can generate data from a wide-ranging variety of distributions. | - Can be difficult to train,<br>- Can be computationally expensive. |

## 4. Opportunities of GenAI in Construction

### 4.1. Current GenAI Applications and Developments in Construction

Recent studies using LLMs to solve construction-related problems demonstrate the long-term opportunities of GenAI in the industry. In 2023, Zheng and Fischer developed a BIM-GPT integrated framework [97] to retrieve, summarize, and answer questions from the BIM database, overcoming the challenges due to the extensive engineering required to automate complex information extraction from rich BIM models. By prompting the LLM appropriately, BIM-GPT shows how advanced integration can extract value from construction data assets. In the early days, such a pioneering idea laid the groundwork for GenAI in the AEC domain. A recent work by Prieto et al. in 2023 [98] shows the potential for large language models to automate repetitive, time-intensive construction tasks. Their study tested using ChatGPT to generate coherent schedules that logically sequence activities and meet scope requirements. Hasan et al. proposed a novel method for classifying injury narratives to identify risks and hazards in construction by fine-tuning bidirectional encoder representations from transformers (BERT) sentence-pair models [99]. The BERT-based approach was also utilized for the automatic detection of contractual risk clauses within construction specifications [100]. A study indicated that limited language generation applications in construction despite extensive documentation such as drawings, reports, and contract documents, cannot feed intelligent systems, though they contain critical references for decisions. Generative AI-like technologies such as ChatGPT and BARD can enable automated synthesis of construction documents and question answering, overcoming analog barriers to unlock the value in this data [101]. In construction automation, the major challenge in maximizing robotic systems is creating efficient sequence planning for construction tasks. Current methods, including mathematics, and machine learning, have limitations in adapting to dynamic construction settings. To address this, a recent study introduced RoboGPT, leveraging ChatGPT's advanced reasoning for automated sequence planning in robot-based construction assembly [102]. The recent CREATE AI Act authorizing the National Artificial Intelligence Research Resource (NAIRR) indicates growing government interest in expanding AI development. By providing open access to key AI resources, NAIRR aims to catalyze innovation across sectors while also serving as a testbed for trustworthy AI practices. Though in the early stages, this initiative represents an important step toward equitable AI advancement by connecting public infrastructure to circulate capabilities more widely through academia and industry[103].

Given the rapid development and deployment of LLMs in recent years, comparing LLMs is useful for tracking progress in this fast-moving field and understanding tradeoffs between model scale, and accessibility to provide an at-a-glance overview for researchers and practitioners. The training parameter size indicates the scale and potential capability of LLMs, giving users insight into model strength, and infrastructure requirements. Bigger models with more parameters tend to be more powerful, generally costlier and need more computational resources.

The LLMs include both open-source and closed-source approaches, each with distinct implications for access, innovation, and collective development. On one hand, open-source large language models promote transparency by providing public access to critical model assets like source code, training data, and model parameters. With freely available implementation details, open source fosters collaboration as developers and researchers can contribute to enhancing and customizing the models to align with specific needs. However, hosting and maintaining accessible open-source models incur infrastructure costs. In contrast, closed-source LLMs are proprietary models restricted to license-holder organizations. Without access to the underlying code, the specific details of the architecture, and training data, the algorithms of closed-source LLMs may not be known to the public. While commercial closed-source models may ensure consistent uptime through dedicated cloud resources, their lack of public transparency limits external innovation opportunities. At the same time, closed-source models carry the advantage of preserving training data privacy. Table 2 summarizes the top ten LLMs currently available, and offers insights for developers and researchers to evaluate both open-source and closed-source options against capability, and updated time when selecting a model aligned with their priorities and constraints.

Table 2. Current Ten Largest LLMs [94], [95], [96], [97], [97]–[102]

| | **LLM** | **Developed by** | **Training Parameter Size (Billion)** | **Release Year** | **Access** |
|---|---|---|---|---|---|
| **1** | GPT-4 | OpenAI | 1000+ | 2023 | Closed |
| **2** | PaLM | Google AI | 540 | 2022 | Open |
| **3** | MT-NLG | Nvidia | 530 | 2021 | Closed |
| **4** | Llama 2 | Meta AI | 500 | 2023 | Open |
| **5** | Gopher | DeepMind | 280 | 2021 | Open |
| **6** | GPT-3.5 | OpenAI | 175 | 2022 | Closed |
| **7** | GPT-3 | Open AI | 175 | 2020 | Closed |
| **8** | OPT | Meta AI | 175 | 2022 | Open |
| **9** | LaMDA | Google AI | 137 | 2022 | Open |
| **10** | GPT-NeoX | Microsoft | 100 | 2023 | Closed |

### 4.2. What Opportunities are Perceived by Construction Industry Practitioners?

To gain insights into construction industry professionals' perspectives on GenAI, various text analytics techniques were applied. A word cloud uncovered frequent key terms, sentiment analysis indicated overall sentiment, and opportunities list synthesized potential application areas. This comprehensive text data analysis provides a picture of discussion topics, attitudes, and outlooks regarding the potential of integrating GenAI into the construction industry.

A word cloud visualization of the LinkedIn data provides an overview of frequently mentioned terms related to generative AI in construction (Figure 2). A word cloud provides a visual representation of textual data, serving as an impactful tool for text analysis [113], [114]. We preprocessed the data by cleaning and tokenization to improve quality. Text cleaning involved formatting adjustments to improve computational readability. Tokenization segmented the text into discrete, meaningful units by isolating individual words and phrases. We then utilized the Natural Language Toolkit (NLTK) in Python to remove generic stop words and distill the corpus down to substantive terms [115], [116]. This shaped a refined dataset with reduced noise, ready for analysis. The results summarize a diverse range of terms that capture the overarching themes and trends within the dataset. The most dominant word is "ai" highlighting the increased attention on artificial intelligence technologies broadly. Notably, "generative" appears with high frequencies demonstrating awareness of this specific AI subdomain. Other common terms like "design", "data", "project", and "technology" indicate a focus on potential applications in construction processes. "ChatGPT" arises fairly often as well, suggesting this popular demo has significantly shaped industry impressions of generative AI capabilities and potential applications in construction. Numerous terms point to opportunities like "productivity", "designs" "tools", and "processes". Meanwhile, words such as "help", "need", "could", and "future" convey a sense of anticipation and speculation around GenAI's developing impacts. Taken together, the word cloud provides a snapshot of how construction professionals are engaging with the emergent GenAI phenomenon, highlighting key opportunities while also indicating uncertainty about optimal applications and next steps.

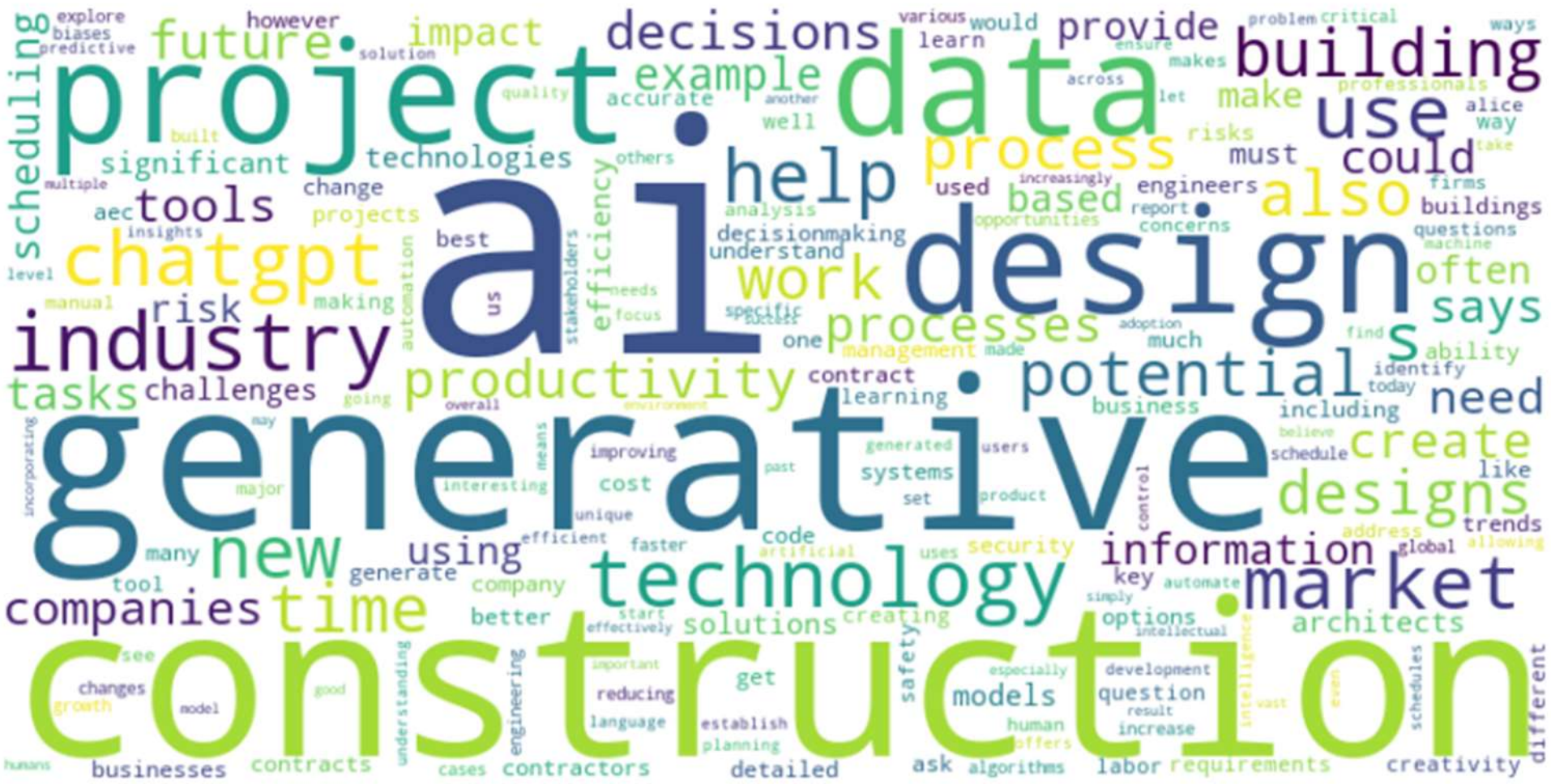

Figure 3. Word Cloud Analysis of Industry Practitioners' Opinions

Furthermore, it is important to uncover the underlying sentiments conveyed in the text. Sentiment analysis, also called opinion mining, involves using computational methods to determine the opinions, attitudes, and emotions expressed toward a subject[114], [117], [118]. Sentiment analysis classifies opinions at three levels: document level categorizes the sentiment of entire documents; sentence-level determines the sentiment of each sentence; and aspect-level examines deeper to categorize sentiment towards specific entity aspects[119]. In our study, we utilized the TextBlob library to quantify sentiment polarity scores, ranging from -1 to 1, revealing positive, negative, or neutral sentiment. Through preprocessing, tokenization, and model-driven analysis, we categorized each text segment. In our sentiment analysis, the discernment of emotional tonality yielded a remarkable distribution: a predominant positivity, coupled with very small negativity and an equivalent neutrality. This outcome highlights the overwhelmingly positive sentiment inherent within the analyzed corpus about GenAI in construction. Visualization using a bar chart showed proportions of positive, negative, and neutral sentiments as shown in Figure 4.

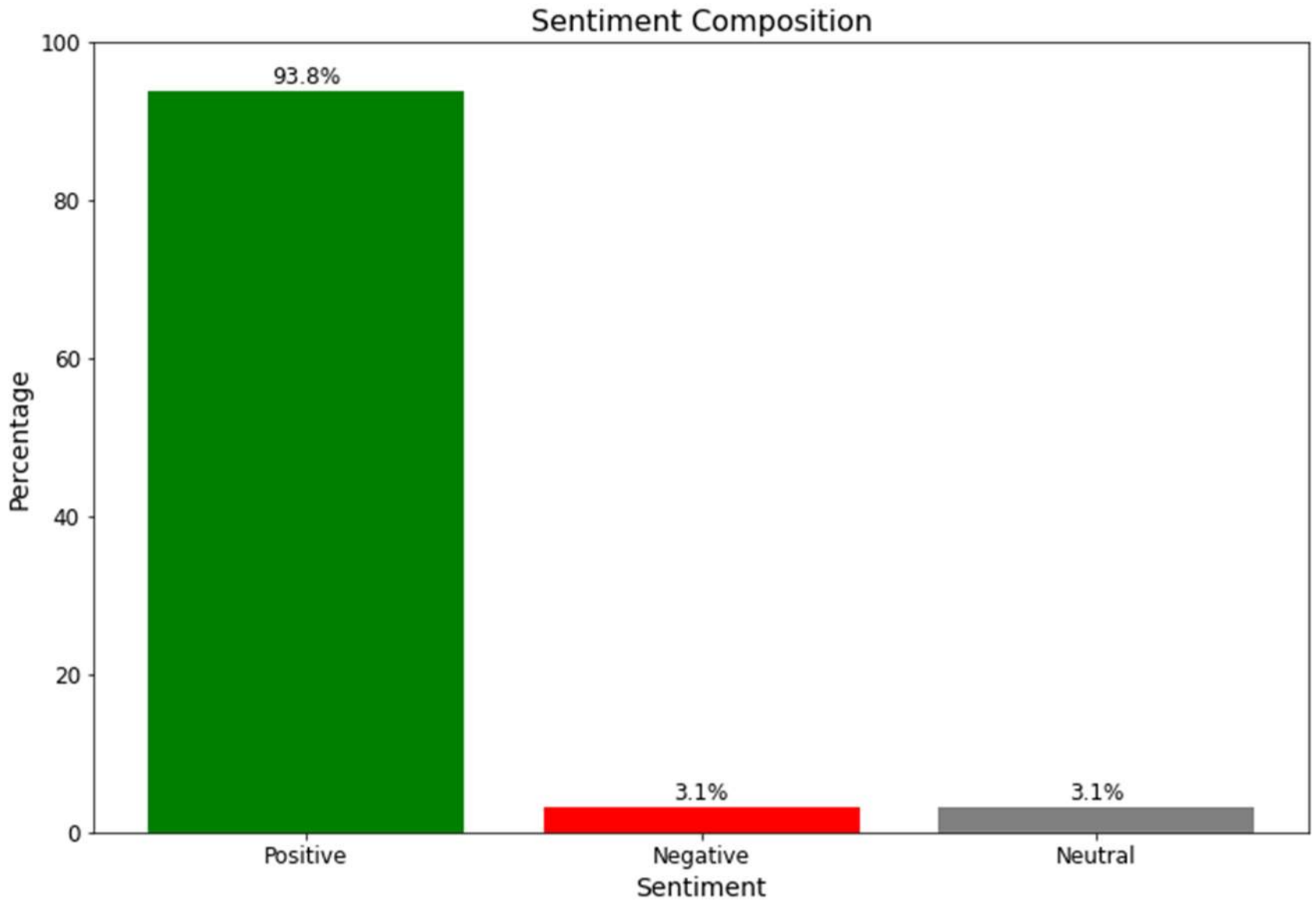

Figure 4. Sentiment Analysis of Industry Practitioners' Opinions

Based on the analysis of people's perspectives, this study synthesizes the key themes regarding the potential opportunities of Generative AI in construction as mentioned in Table 3. First, we identified the main points and common ideas expressed across multiple perspectives in the body of the text through careful reading and analysis. Second, we synthesized these main points into a few key, overarching themes that capture the essence of the perspectives. There is consensus around Generative AI's promise to drive greater efficiency, innovation, and data-driven decision-making across the construction lifecycle. However, viewpoints diverge regarding the scale and scope of GenAI's applications, as well as the need to thoughtfully manage its integration to maximize benefits versus risks.

Table 3. Overarching Themes on Opportunities

| Perspectives: Main Points | Key Theme |
|---|---|
| Applying GenAI for construction documents management | **Construction Documents and Data Management** |
| Enterprise search | |
| Data management ultimately offers time-saving benefits and increased productivity when effectively leveraged | |
| For example, Integrating GenAI in scheduling to identify the most effective schedule path to follow. | |
| Can help improve conversations and collaboration between project stakeholders such as contractors, designers, and owners. | **Question Answering (QnA)**: |
| Stakeholder demands for faster, affordable, and sustainable builds create opportunities for GenAI and automation to address construction's unique challenges such as repetitive tasks and unsafe work environments. | **Automation for Unique Challenges** |
| AI-generated designs and plans reduce manual work, enhancing data systems for faster payments, fewer errors, and better decisions. | **AI-Generated Designs** |
| Generative AI increases predictive capabilities, leveraging historical data for accurate project forecasting, forecasting of trends, risk assessment, and opportunity identification. | **Accurate Forecasting** |
| Incorporating GenAI streamlines the synthesis of project data and provides avenues for automating intricate information management, such as contract-related data, thereby enhancing decision-making during the initial phases of construction. | **Project Data Synthesis** |
| AI and modern innovations in construction address labor shortages, cost escalation, and environmental concerns, positioning the industry for a transformative future. | **Efficiency and Sustainability** |
| Integrate materials assessment AI tools to support informed materials selection for improved sustainability, maximizing de-carbonization. | **Materials Assessment** |
| The development of GenAI, like ChatGPT, enhances human capabilities rather than replacing jobs. | **AI Augmentation** |

### 4.3. Potential Applications of GenAI in Construction

Generative AI shows huge potential to transform information workflows in architecture, engineering, and construction. Advanced LLMs can parse volumes of unstructured data to extract insights with new levels of ease and speed. For instance, by analyzing building codes, generative models can identify relevant requirements and produce summarized, project-specific reports for architects. This automates laborious manual reviews. Similarly, contractors can input design specifications into AI systems to automatically compile cost and schedule estimates by associating 3D models with external databases. Many simple properties like material name, soil type, concrete strength, roof slope, furniture suppliers, last changed by, as well as complex analytical queries become accessible to stakeholders through AI's natural language capabilities. Whether generating code requirements from regulations, connecting designs to cost data, or retrieving wind load assumptions, GenAI allows seamless information flow between physical and virtual manifestations of the built environment. The power of language models lies in their ability to comprehend, reason about, and generate knowledge. As explained through these use cases, GenAI can improve project understanding and decision-making by unlocking information trapped in unstructured data. The GenAI holds vast potential to increase productivity and collaboration in the AEC industry.

In this section, based on lessons learned from literature, peoples' perspectives, and building lifecycle tasks identified[120]–[123], we provide the potential application examples across the project lifecycle, detailing beneficiaries and appropriate GenAI model types for each as shown in Table 4. Clearly defining the output modality generated by each AI system, whether text, image, 3D, video, or task, simplifies technical requirements for implementation. Readers can identify suitable architectures by mapping desired functionality to output types. In addition, clustering potential applications by common model families also enables knowledge transfer across use cases and highlights productive pairings of activities with generative techniques. In addition, the popular model examples of each type at the end of the table expedites the process of model selection, allowing researchers and practitioners to make quicker decisions customized to their specific application requirements and objectives.

Table 4. Potential Applications of GenAI in Different Phases of Building Lifecycle

| Phase | Potential GenAI Application | Main beneficiary | Model type based on the output |
|---|---|---|---|
| Feasibility | • To generate a feasibility report | Stakeholders | text-to-text |
| | • To generate a project initiation document (PID) | Owner | text-to-text |
| | • Interactive Q&A chatbot to refine PID details | Owner | text-to-text |
| | • To create visual representations of data such as site conditions, traffic patterns, zoning laws, etc. | Stakeholders | text-to-image |
| | • To predict project milestones and success criteria for different phases of the project | Stakeholders | text-to-text |
| | • To create contracts and agreements | Stakeholders | text-to-text |
| Design | • To generate multiple conceptual designs based on the program requirements and communicate with the architect | Architect | text-to-task |
| | • Animated 3D visualization of organization chart and responsibilities | Stakeholders | text-to-3D |
| | • To automatically generate detailed cost estimation report | Owner | text-to-text |
| | • To associate cost/time data with building design | Contractor | text-to-text |
| | • To extract structural design requirements | Engineer | text-to-text |
| | • To extract MEP design requirements | Engineers | text-to-text |
| | • To generate a permit application draft | Architect | text-to-text |
| | • To generate a risk analysis report | Stakeholders | text-to-text |
| | • To develop a design communication Q&A chatbot | Architect | text-to-text |
| | • To compare the design against the building code requirements | Architect | text-to-task |
| | • To perform complex design checking (routing analysis, etc.) | Architect | text-to-task |
| | • To select the most suitable contractors based on project-specific criteria, performance histories, and contractual considerations | Owner | text-to-text |
| Procurement | • To visualize the material delivery schedule | Logistics team | text-to-3D |
| | • To generate a request for a quotation | Procurement team | text-to-text |
| | • Identification of optimal supplier based on variables | Project manager | text-to-text |
| | • Streamline subcontractor bidding and selection | Contractor | text-to-text |
| | • Automated inventory management | Procurement team | text-to-text |
| Construction | • To extract project information from construction documents such as dimensions, materials used, responsible person, point of contact, etc. | Contractor | text-to-text |
| | • To generate new documents. Examples- proposals, reports, etc. | Contractor | text-to-text |
| | • To classify & cluster documents based on project types, internal departments, sub-contractors, project phases, document types, materials, supply chain, etc. | Contractor | text-to-text |
| | • Generating code to automate tasks. | Contractor | text-to-text |
| | • Translating documents into different languages. | Contractor | text-to-text |
| | • To optimize cost estimation workflow | Estimator | text-to-task |
| | • To help progress tracking and identify safety concerns with drone integration | | text-to-task |
| | • To provide customized alerts and notifications on changes | | text-to-task |
| | • To help quality control such as comparing completed tasks to project specifications to identify defects and deviations | | text-to-text |
| | • To generate an optimal schedule path | Contractor | text-to-text |
| | • Searching specific information on the data lake, shared folder, project-specific information, etc. | Contractor | text-to-task |
| | • To generate targeted safety training materials | Safety manager | text-to-image/text-to-video |
| | • To generate targeted trade training materials | Project manager | text-to-image/text-to-video |
| Operation & Maintenance | • To create a knowledge management system using a Q&A chatbot | Facility manager | text-to-text |
| | • To create a work order from logs | Technician | text-to-text |
| | • Generative design of replacement parts | Technician | text-to-3D |
| | • To generate maintenance schedule & predictive maintenance | Facility manager | text-to-text/text-to-image |
| | • To generate an energy consumption report | Facility manager | text-to-text |
| | • Chabot to assist occupants | Occupants | text-to-text |
| Any Phase | • Information retrieval from nD BIM models | Stakeholders | text-to-text/text-to-3D |
| Any Phase | • Updating the nD BIM model using natural language commands | Stakeholders | text-to-task/text-to-3D |
| Any Phase | • Robot instruction in natural language | Stakeholders | text-to-task |
| Any Phase | • To help human-robot interaction | Stakeholders | text-to-task |
| **Some useful model examples based on their output:** | | | |
| *Text-to-text models: ChatGPT, LaMDA, PEER, Galactica, Codex, Claude, Jurassic*<br>*Text-to-image models: DALL-E2, Parti, IMAGEN, Craiyon*<br>*Text-to-video/3D models: Imagen, StyleGAN, Phenaki, Magic3D*<br>*Text-to-task models: Bard, GPT-4, LaMDA, Jarvis* | | | |

### 4.4. A Conceptual Implementation Framework

To accomplish identified potential applications in reality, this study presents a conceptual GenAI implementation framework in construction for early adoption in research and industry application. The framework for fine-tuning generative LLM comprises three interconnected stages: selection, fine-tuning, and utilization as shown in Figure 5. First, a potential application is identified based on the requirements and objectives. Next, a model type that aligns with the desired output and objectives is determined, and then a base LLM is selected from providers like OpenAI, Meta, etc., that leverage diverse knowledge sources and model architectures (e.g. GPT-3, Llama). The model may be open source with available code and data, or proprietary with just API access. Next, domain-specific data is collected for fine-tuning, such as BIM data, cloud-based data repositories, and various other datasets in construction. Fine-tuning involves techniques such as parameter adjustment and rewards to align the model with the desired objectives, which may include privacy constraints and noise reduction to enhance the model's performance. The resulting fine-tuned model has knowledge customized for the construction domain. Finally, the adapted model is deployed through careful prompt engineering to query its capabilities. Users provide prompts and obtain answers or visualizations based on the fine-tuned model's specialized intelligence. This conceptual framework for fine-tuning LLMs bridges the gap between pre-trained models and enterprise-specific applications, promoting adaptability in a wide range of domains.

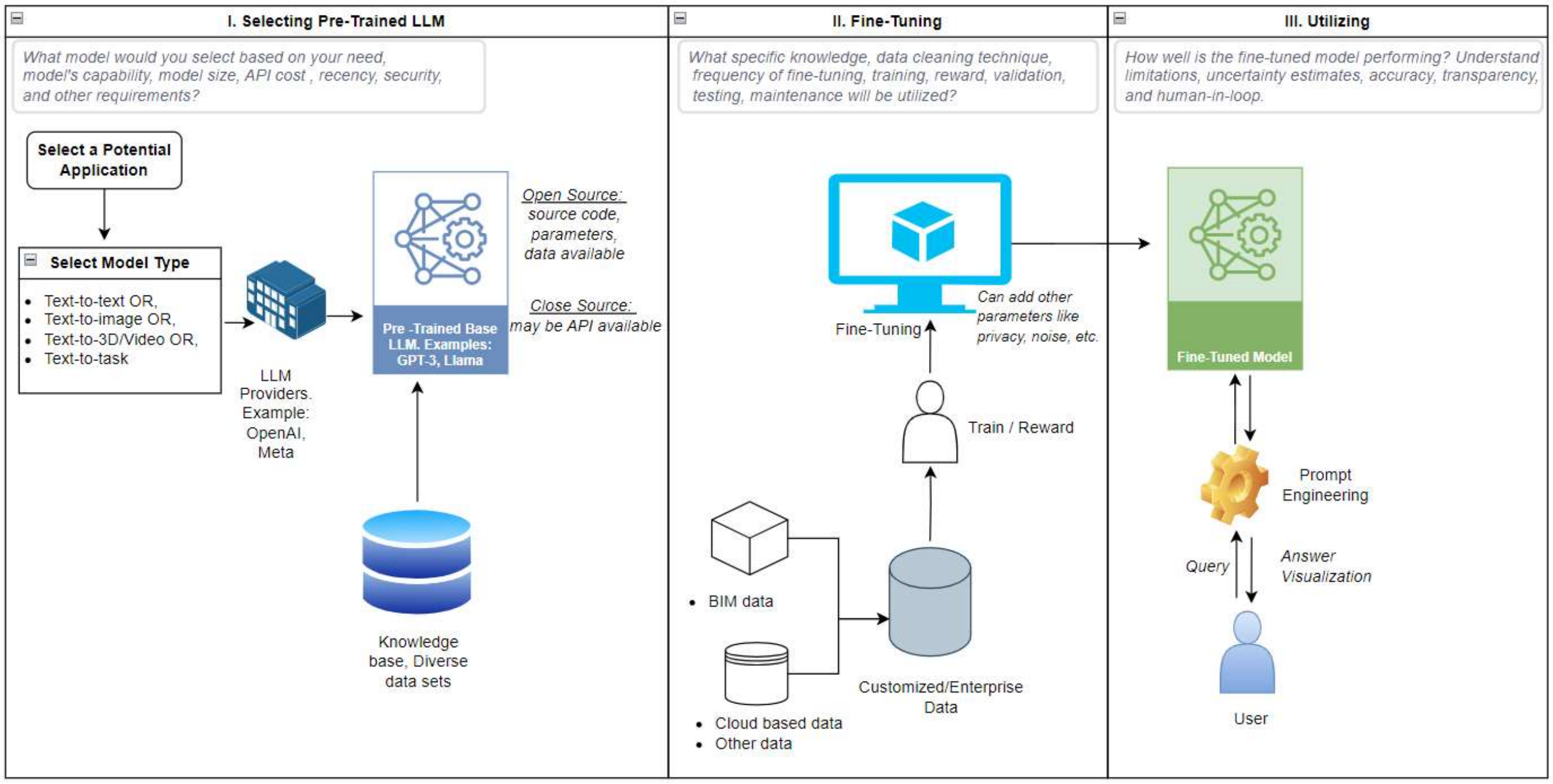

Figure 5. A Conceptual GenAI Implementation Framework

## 5. Challenges of GenAI Implementation in Construction

Generative AI adoption across industries is rapidly growing, driven by the immediate integration of new technologies like ChatGPT intensifying competitive pressures on organizations while this novelty presents new risks[69]. Like other industries, the integration of GenAI in construction is associated with complex challenges. Therefore, it is important to understand these challenges before applying the proposed conceptual framework. These challenges comprise various areas, including domain knowledge, the potential for hallucinations in AI-generated outputs, the crucial aspect of accuracy in AI predictions, the generalizability of AI models to new situations, the need for frequent model updates and interpretability, the cost implications of deploying generative AI, and the ethical considerations around data privacy, bias, and accountability as shown in Figure 6. Furthermore, the construction sector faces specific regulatory hurdles related to the responsible use of GenAI, prompting the need for AI skill development and training, liability determination, copyright and intellectual property concerns, and certification protocols. Addressing these multidimensional challenges requires a proactive and collaborative effort involving industry experts, policymakers, and AI researchers to ensure the safe and effective implementation of GenAI in construction practices.

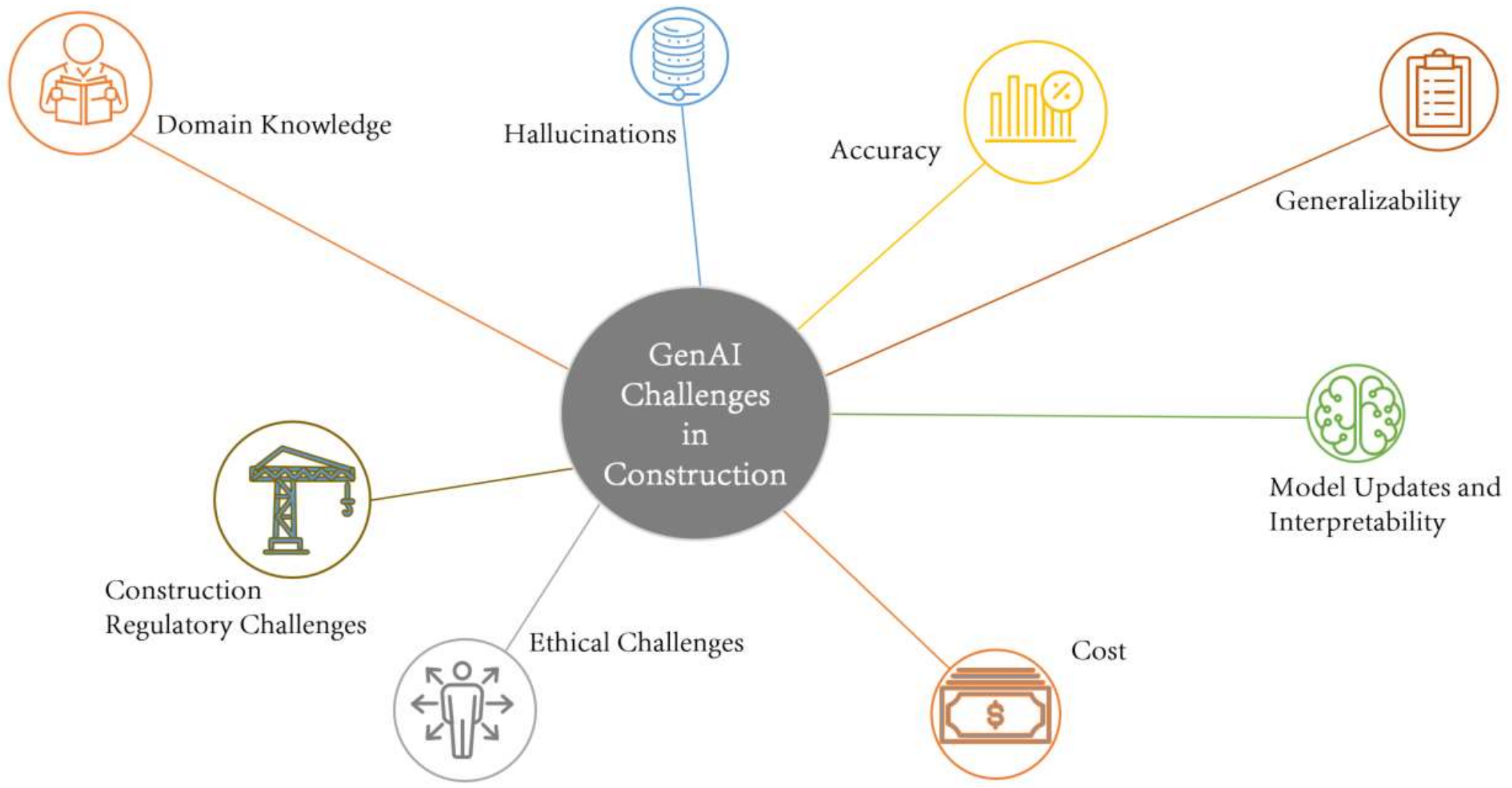

Figure 6. Challenges of GenAI in Construction

### 5.1. Domain knowledge

The construction industry poses unique difficulties in applying GenAI due to its vast domain knowledge requirements. Capturing the industry's complicated technical engineering expertise across structural, mechanical, electrical, plumbing, and project management disciplines remains challenging. Construction also relies heavily on physical situational awareness and spatial reasoning when manipulating materials and navigating dynamic job site capabilities stretching the limits of AI [36]. Consequently, construction's vast knowledge context hinders GenAI's ability to extract meaningful structure-activity relationships from industry data. However, promising avenues exist to address these knowledge gaps. For instance, large language models like GPT require fine-tuning and contextual input tailored to the construction domain in order to efficiently generate industry-specific insights [124]. Hybrid reasoning techniques combining top-down ontological, symbolic knowledge with bottom-up neural networks can be beneficial. Therefore, advancing construction-focused GenAI requires incorporating domain knowledge more seamlessly into model architecture and training. This domain knowledge infusion remains an open research area for unlocking GenAI that can meet construction's complex and ever-changing demands.

### 5.2. Hallucinations

Generative artificial intelligence systems face challenges with hallucination, generating convincing but false outputs due to limited knowledge [70]. These hallucinations often result from factors such as inadequate or noisy training data, a lack of contextual understanding, or imposed constraints. GenAI systems are particularly notorious for producing aesthetically pleasing yet inaccurate predictions, often with an unwarranted high level of confidence. For instance, in the context of a GenAI scheduling system, hallucinations could lead to the generation of inaccurate timelines for critical paths. In construction-focused AI, which lacks the capability to perceive and validate real-world complexities directly, there is a risk of generating hallucinatory outputs that are apart from reality. To mitigate these potentially unsafe hallucinations, several strategies can be employed. These include the use of high-quality training data, a strong grounding in engineering and construction knowledge, simulated testing to validate predictions, continuous monitoring of uncertainty, and the introduction of human oversight throughout the AI's decision-making processes.

## 5.3. Accuracy

Ensuring accuracy is a major challenge for GenAI, as inappropriate outputs can lead to big failures. Large language models like GPT-3 show these limits, relying on minimal training data from unverified sources [125]. Lack of fundamental construction engineering knowledge, such models obtain only superficial statistical associations rather than causal basics, risking construction decisions through misguided outputs. However, techniques exist to enhance output validity. Construction-specific fine-tuning with validated datasets can align models to the complexities of the built environment. Uncertainty indicators can flag doubtful predictions needing additional verification. Simulated testing enables early correction of inaccuracies before real-world implementation [126]. Further, prompted self-improvement may allow models to iteratively refine their outputs [127]. Overall, connecting robust datasets, uncertainty metrics, simulated validation, and self-correction procedures can introduce proper engineering causality over statistics, improving construction GenAI's accuracy. Advancing fundamental reasoning capabilities remains critical for developing generative intelligent systems that meet the construction industry's need for reliable automation and decision-making.

## 5.4. Generalizability

Generalizability refers to the ability of a generative AI model to extend its learning beyond the specific datasets and distributions it was trained on. A GenAI system utilizing historical data may encounter issues with poor generalization, where the knowledge derived from training data in the in-sample period does not effectively apply to new, out-of-sample data in testing. Even if a model fits the training data well, its poor generalization is unusable for addressing real-world decision-making challenges [128]. For example, a model pre-trained on fixed historical data may fail to account for unexpected changes like weather delays, labor availability, or design changes. Models trained on a limited dataset, unfamiliar inputs, and lack of a casual understanding mechanism in the model are the major challenges that contribute to the generalizability problem. Collecting diverse training data and testing models on novel inputs helps the construction GenAI better generalize [129]. Leveraging simulation, causal reasoning, and common-sense checks also improves generalization by teaching strong process knowledge. And, continual learning enables adaptation to new data over time. Together these solutions improve generalization.

## 5.5. Model Updates and Interpretability

Model updating is a key challenge for deploying generative AI in construction. Training data can quickly become outdated as materials, methods, and regulations frequently change. Without recent data, models will miss new innovations and provide unreliable guidance. For example, an AI chatbot trained before the pandemic may overlook the impacts of supply chain disruptions and labor shortages. Regularly retraining models on new data is essential, but costly and complex at scale. Potential solutions include modular model architectures to simplify updating, simulations to generate fresh synthetic training data, and lightweight model adaptation techniques like transfer learning. However, balancing model accuracy and update will remain an obstacle. User oversight and paired human-AI collaboration are recommended when utilizing construction generative AI. In addition, another limitation of deep generative models is their black-box nature - the internal workings are not transparent or easily interpretable. This is problematic for critical construction applications where explainability is important [130], [131]. The opaque processes by which generative AI systems produce outputs create uncertainties around reliability and trustworthiness. Users cannot validate which parts of the model's knowledge base are being leveraged. Therefore, more research is needed to develop interpretable model architectures and training techniques, making the decision-making logic clear. Progress in the construction of explainable AI will be key to wider adoption by explaining the reasoning behind outputs and establishing confidence in the technology.

## 5.6. Cost

Training and operating generative AI models require significant costs, presenting challenges for widespread construction industry adoption. The training phase alone demands massive computing resources and time to produce capable generative capacity. Ongoing operating expenses also accumulate from the energy required to run large models and web-serving infrastructure [2]. For example, monthly subscription fees to access ChatGPT currently start at $20 with traffic limitations. In addition, utilizing GPT models to develop conversational apps produces additional usage costs billed per generated token [124]. Initial application development leveraging these models is expensive upfront too. The considerable resource demands and ongoing costs act as barriers, especially for smaller construction companies with limited budgets [132]. Further optimizations to reduce the computing power, energy, and data needs of generative models would support feasibility. More cost-effective scaling solutions tailored for construction use cases could also expand access. Overcoming these cost challenges requires a well-balanced approach, considering the long-term benefits of GenAI integration against the upfront investments needed to tie together its capabilities effectively.

## 5.7. Ethical Challenges

The adoption of generative AI models also raises ethical issues around data privacy, bias, and accountability that the construction industry must proactively address. These data-intensive models can utilize sensitive project information and personal details lacking proper consent, presenting risks of confidentiality breaches and intellectual property violations. Researchers and the industry should implement data privacy safeguards and anonymization measures. For example, OpenAI's ChatGPT explicitly acknowledges its potential to generate inaccurate information about individuals, locations, or facts, underlining the need for researchers to be aware of this limitation and ethical challenges when incorporating ChatGPT in scientific works. This includes essential considerations regarding data privacy, confidentiality, and informed consent [133]. The handling of sensitive data by ChatGPT introduces vulnerabilities that may be exploited for unauthorized access or misuse, thereby posing substantial privacy and security risks [69]. Also, the adoption of LLMs raises concerns about creating potential biases[134]. The utilization of confidential construction data like cost, schedule, safety records, contract documents, and BIM model information may potentially trespass upon intellectual property rights and give rise to ethical and legal difficulties. Therefore, establishing clear accountability for errors or accidents caused by AI-generated outputs remains a complex issue needing careful consideration, in order to develop ethically responsible frameworks for implementing generative AI within the construction industry.

## 5.8. Construction Regulatory Challenges

In the construction sector, the integration of GenAI poses several complex regulatory challenges. Successful implementation requires AI understanding, skillsets, and trainings so that industry experts can properly utilize these models. One of the major skills required is proficiency in "prompt engineering," optimizing prompts to maximize model efficacy [124], [135]. However, overreliance on automation risks in reduction of human expertise and the potential for errors in cases of AI malfunction or erroneous information provision [136]. As generative models become capable of autonomously producing comprehensive deliverables, for example, a detailed site safety plan, a serious concern emerges regarding accountability in the event of a failure. Determining liability in such instances, wherein something goes wrong, becomes a complex matter. Who bears responsibility in the event of a failure - is it the developer of the AI system, the construction company implementing it, or the safety manager who approved the final AI-generated plans? Additionally, the independent origination of new content by AI raises questions about copyrights and intellectual property. The ownership of AI-generated content requires a clear legislative definition. To maintain expertise and safety standards, construction companies could introduce certification protocols for AI training and deployment. Moreover, close cooperation between industry experts, policymakers, and AI researchers is essential to navigate these regulatory challenges.

### 5.9. What Challenges are Perceived by Construction Industry Practitioners?

The challenges obstructing GenAI adoption in construction are associated with both technological and human factors. A recent LinkedIn poll of 48 AEC professionals investigated the frequency of generative AI usage in their work, finding 40% have never tried it, 33% use it sometimes, 19% use it often, and 8% use it all the time[137]. This reveals that most AEC professionals are still in the early stages of generative AI adoption, though a segment has integrated these tools into their regular workflows. And, another poll of 16 AEC professionals examined whether their organizations have policies regarding the use of commercial GenAI tools, finding 63% do not, 31% do, and 6% are unsure[137]. This indicates that most companies currently lack formal guidelines on GenAI usage, presenting an opportunity to implement policies and controls given the rise of technologies like ChatGPT. The analysis of perspectives shows key themes around security, governance, awareness, and adaptation as mentioned below. Construction companies must proactively address these multifaceted challenges to unlock their potential. This requires strategic approaches customized to the construction industry's distinct needs within this rapid innovation. A thoughtful, industry-centered path can help overcome obstacles and realize GenAI's potential.

- **Proactive Approach Needed**: The implementation of GenAI in construction requires a proactive approach to security and governance. Addressing these challenges is vital to unlock the potential for improved productivity and creativity during the industry's technological transformation.
- **Strategic Adoption:** The adoption of GenAI within construction companies requires a strategic approach to manage security, risks, and governance effectively. The practical procedures allow responsible and ethical utilization while maintaining standards of security, safety, and compliance. The guidance from construction technology experts can support in setting up a successful generative AI program.
- **Implementation Challenges**: GenAI systems help a comprehensive analysis of trade-offs in construction projects, including physical, financial, and sustainable aspects. However, addressing implementation challenges, such as increasing awareness and understanding, is essential to drive broader adoption and establish convincing business cases for technology investments.
- **Limited Awareness**: The construction industry is facing difficulties in building an efficient business case for investments in software, hardware, training, and infrastructure due to limited awareness. These challenges related to accessing and sharing big data hinder the effectiveness of GenAI models. Moreover, regulatory and legal complexities, particularly concerning intellectual property rights, add compliance concerns when deploying GenAI in visualizations or renderings.
- **Expectation of Mature Technologies**: The construction market expects mature technologies ready for immediate use, focusing on solutions designed to the industry's distinctive challenges. However, this expectation leads to a deeper exploration of automation and AI in construction, recognizing the need for specialized solutions.

- **Risk Mitigation and Ethical Governance:** To effectively implement GenAI in the construction industry, it is important to apply comprehensive risk mitigation strategies. These include various measures such as data encryption, strict access controls, and secure data storage practices. Furthermore, to safeguard AI-generated outcomes, addressing intellectual property concerns through well-defined guidelines and contractual agreements is essential.
- **Novelty Challenge**: Another challenge in applying GenAI lies in its novelty. For example, many traditional schedulers are familiar with long-standing tools and may hesitate to embrace newer, more advanced solutions.

## 6. Recommendations and Future Directions

In section 4.3, we have explained various potential applications that serve as a foundation for future research directions. We have structured this section into two subsections: 1) recommendations: short-term and long-term adaption strategies and, 2) future research directions: major future research questions. These sections show the directions for studies aimed at facilitating the effective integration of GenAI within the industry.

### 6.1. Recommendations

We recommend the following short-term and long-term strategies for adapting GenAI in construction:

- **Fine Tuning LLMs:** The recommended initial approach for the integration of GenAI into the construction industry involves the fine-tuning of available powerful pre-trained language models using construction-specific data. Construction companies have the opportunity to curate datasets comprising various resources such as design documents, building codes, contractual documents, technical documents, and BIM data. This data is helpful in informing the selected LLM about specialized vocabulary and contextual nuances of the construction. Starting with modest datasets and focusing on strongly defined tasks can simplify the process of prompt engineering that enables the GenAI systems for construction needs.
- **Human Oversight:** GenAI systems still require human oversight to validate quality and accuracy while capable of automating tasks. Model outputs should be reviewed and feedback can be provided to improve performance. Therefore, human-in-the-loop approaches that combine AI generation with human judgment can improve the strengths of both.
- **Evaluating Business Impact:** It is recommended to assess the business impacts of GenAI using experiments measuring key performance indicators. Pilot studies could evaluate model influence on metrics such as productivity, cost, time, risks, etc. The measurement as a model integrates more data and provides insight into returns over investment. This can help to quantify the benefits of GenAI investment for the organization.
- **Developing Custom LLMs:** In the long run, collaborative efforts between the AEC industry and researchers can focus on designing specialized language model architectures for construction-related tasks. This involves compiling extensive datasets from the AEC domain. The fundamental approach is to establish a secure central data repository, with contributions from construction companies, and consultants. Training models on this data, with the support of AI researchers, will allow domain expertise and innovation.

### 6.2. Future Research Directions

We present the following major future research questions for adapting GenAI in construction:

- How can we develop GenAI models that can accurately extract detailed project information from a variety of construction documents and BIM models? This could help improve productivity.
- What techniques can enable GenAI models to automatically generate feasible building designs based on requirements? Generative design could help with time and cost savings.
- How can we build AI assistants that can have natural conversations with human stakeholders to refine project details, requirements, and reports in different phases of the building lifecycle? Conversational AI could help project stakeholders.
- What GenAI techniques can enable the automated generation of 3D visualizations, videos, and images from text descriptions? This could help in better communication.
- How can we develop AI systems to accurately evaluate construction progress, safety, and quality using visual data? Computer vision integration could be key to achieving this.
- What GenAI techniques can optimize construction scheduling, logistics, and cost estimating? This could help in construction project management.
- How can we build AI assistants that can understand BIM model information, extract that information, and update BIM models based on prompts? This could help to accelerate the BIM execution process for general contractors.
- How can we integrate robotics with natural language AI to enable easy human-robot interactions? This could help enhance the usability, and accessibility of robotic systems, leading to improved collaboration.
- What machine learning techniques can support accurate automatic code generation for construction tasks and changes in scope? This could help to track changes and troubleshoot issues.
- How can we build GenAI models that learn continuously from construction data to improve predictions and decision-making over time? This could help in the overall success of an organization, and future project forecasting.

## 7. Conclusion

This study makes important contributions by investigating the evolving opportunities and challenges of implementing Generative AI in the construction industry. Through a detailed literature review, we have identified the limitations of traditional AI methods and examined the recent use cases of GenAI models. We have also investigated the industry practitioners' insights, using sentiment analysis and theme-based interpretation, into the perceived application potential and barriers to adopting GenAI in the construction sector. Synthesizing these findings, we identified potential applications and proposed a conceptual framework to guide researchers and practitioners in implementing GenAI in construction. The mapping of different GenAI model types to various construction tasks suggested potential future applications of text-to-text, text-to-image, text-to-3D/Video, and text-to-task models for applications across project feasibility, design, procurement, construction, and operation phases. However, our study also highlights significant GenAI implementation challenges around domain knowledge, hallucinations, model accuracy, generalizability, interpretability, cost, ethical, and regulatory challenges that must be addressed before executing the proposed framework. Recommendations provided in this study are expected to help construction stakeholders with strategies for initiating GenAI adoption and plan for long-term application while mitigating risks. The future research questions identified can direct the construction research community to focus on the practical applications of GenAI capabilities. Moreover, this study provides a strong literature foundation for realizing the capacity and challenges of GenAI in this industry. Further validation studies implementing the proposed framework and developing real construction applications would be a natural extension of this research.

**Funding**

This research received no external funding.

**Conflicts of Interest**

The authors declare no conflict of interest.

**References**

[1] Y. Xu, Y. Zhou, P. Sekula, and L. Ding, "Machine learning in construction: From shallow to deep learning," *Developments in the Built Environment*, vol. 6, p. 100045, May 2021, doi: 10.1016/j.dibe.2021.100045.
[2] Y. Liu *et al.*, "Generative artificial intelligence and its applications in materials science: Current situation and future perspectives," *Journal of Materiomics*, vol. 9, no. 4, pp. 798–816, Jul. 2023, doi: 10.1016/j.jmat.2023.05.001.
[3] Y. Duan, J. S. Edwards, and Y. K. Dwivedi, "Artificial intelligence for decision making in the era of Big Data – evolution, challenges and research agenda," *International Journal of Information Management*, vol. 48, pp. 63–71, Oct. 2019, doi: 10.1016/j.ijinfomgt.2019.01.021.
[4] D. Baidoo-Anu and L. Owusu Ansah, "Education in the Era of Generative Artificial Intelligence (AI): Understanding the Potential Benefits of ChatGPT in Promoting Teaching and Learning." Rochester, NY, Jan. 25, 2023. doi: 10.2139/ssrn.4337484.
[5] Qwiklabs, "Introduction to Generative AI | Google Cloud Skills Boost," *Qwiklabs*. https://www.cloudskillsboost.google/course_sessions/4093050/video/384243 (accessed Aug. 16, 2023).
[6] C. Li, Y. Su, and W. Liu, "Text-To-Text Generative Adversarial Networks," in *2018 International Joint Conference on Neural Networks (IJCNN)*, Jul. 2018, pp. 1–7. doi: 10.1109/IJCNN.2018.8489624.
[7] C. Zhang, C. Zhang, M. Zhang, and I. S. Kweon, "Text-to-image Diffusion Models in Generative AI: A Survey." arXiv, Apr. 02, 2023. Accessed: Aug. 27, 2023. [Online]. Available: http://arxiv.org/abs/2303.07909
[8] V. Liu, T. Long, N. Raw, and L. Chilton, "Generative Disco: Text-to-Video Generation for Music Visualization." arXiv, Apr. 17, 2023. Accessed: Aug. 27, 2023. [Online]. Available: http://arxiv.org/abs/2304.08551
[9] T. Lei, R. Barzilay, and T. Jaakkola, "Rationalizing Neural Predictions." arXiv, Nov. 02, 2016. Accessed: Aug. 27, 2023. [Online]. Available: http://arxiv.org/abs/1606.04155
[10] R. Gozalo-Brizuela and E. C. Garrido-Merchan, "ChatGPT is not all you need. A State of the Art Review of large Generative AI models." arXiv, Jan. 11, 2023. Accessed: Aug. 15, 2023. [Online]. Available: http://arxiv.org/abs/2301.04655
[11] C. Q. X. Poh, C. U. Ubeynarayana, and Y. M. Goh, "Safety leading indicators for construction sites: A machine learning approach," *Automation in Construction*, vol. 93, pp. 375–386, Sep. 2018, doi: 10.1016/j.autcon.2018.03.022.
[12] Y. M. Goh and D. Chua, "Neural network analysis of construction safety management systems: a case study in Singapore," *Construction Management and Economics*, vol. 31, no. 5, pp. 460–470, May 2013, doi: 10.1080/01446193.2013.797095.
[13] D. K. Chua and Y. M. Goh, "Poisson Model of Construction Incident Occurrence," *Journal of Construction Engineering and Management*, vol. 131, no. 6, pp. 715–722, Jun. 2005, doi: 10.1061/(ASCE)0733-9364(2005)131:6(715).

[14] W. Fang *et al.*, “Computer vision applications in construction safety assurance,” *Automation in Construction*, vol. 110, p. 103013, Feb. 2020, doi: 10.1016/j.autcon.2019.103013.
[15] J. Liu, H. Luo, and H. Liu, “Deep learning-based data analytics for safety in construction,” *Automation in Construction*, vol. 140, p. 104302, Aug. 2022, doi: 10.1016/j.autcon.2022.104302.
[16] T. P. Williams and J. Gong, “Predicting construction cost overruns using text mining, numerical data and ensemble classifiers,” *Automation in Construction*, vol. 43, pp. 23–29, Jul. 2014, doi: 10.1016/j.autcon.2014.02.014.
[17] M.-Y. Cheng, H.-C. Tsai, and W.-S. Hsieh, “Web-based conceptual cost estimates for construction projects using Evolutionary Fuzzy Neural Inference Model,” *Automation in Construction*, vol. 18, no. 2, pp. 164–172, Mar. 2009, doi: 10.1016/j.autcon.2008.07.001.
[18] P. Ghimire, S. Pokharel, K. Kim, and P. Barutha, *MACHINE LEARNING-BASED PREDICTION MODELS FOR BUDGET FORECAST IN CAPITAL CONSTRUCTION*. 2023.
[19] S. K. Baduge *et al.*, “Artificial intelligence and smart vision for building and construction 4.0: Machine and deep learning methods and applications,” *Automation in Construction*, vol. 141, p. 104440, Sep. 2022, doi: 10.1016/j.autcon.2022.104440.
[20] A. Mahmoodzadeh, H. R. Nejati, and M. Mohammadi, “Optimized machine learning modelling for predicting the construction cost and duration of tunnelling projects,” *Automation in Construction*, vol. 139, p. 104305, Jul. 2022, doi: 10.1016/j.autcon.2022.104305.
[21] C. Zhang, S. R. Kuppannagari, R. Kannan, and V. K. Prasanna, “Building HVAC Scheduling Using Reinforcement Learning via Neural Network Based Model Approximation,” in *Proceedings of the 6th ACM International Conference on Systems for Energy-Efficient Buildings, Cities, and Transportation*, in BuildSys ’19. New York, NY, USA: Association for Computing Machinery, Nov. 2019, pp. 287–296. doi: 10.1145/3360322.3360861.
[22] S. O. Abioye *et al.*, “Artificial intelligence in the construction industry: A review of present status, opportunities and future challenges,” *Journal of Building Engineering*, vol. 44, p. 103299, Dec. 2021, doi: 10.1016/j.jobe.2021.103299.
[23] J. Chen, Y. Fang, Y. K. Cho, and C. Kim, “Principal Axes Descriptor for Automated Construction-Equipment Classification from Point Clouds,” *Journal of Computing in Civil Engineering*, vol. 31, no. 2, p. 04016058, Mar. 2017, doi: 10.1061/(ASCE)CP.1943-5487.0000628.
[24] S. Sakhakarmi, J. Park, and C. Cho, “Enhanced Machine Learning Classification Accuracy for Scaffolding Safety Using Increased Features,” *Journal of Construction Engineering and Management*, vol. 145, no. 2, p. 04018133, Feb. 2019, doi: 10.1061/(ASCE)CO.1943-7862.0001601.
[25] J. Teizer, “Status quo and open challenges in vision-based sensing and tracking of temporary resources on infrastructure construction sites,” *Advanced Engineering Informatics*, vol. 29, no. 2, pp. 225–238, Apr. 2015, doi: 10.1016/j.aei.2015.03.006.
[26] Z. Zhu and I. Brilakis, “Parameter optimization for automated concrete detection in image data,” *Automation in Construction*, vol. 19, no. 7, pp. 944–953, Nov. 2010, doi: 10.1016/j.autcon.2010.06.008.
[27] F. Pour Rahimian, S. Seyedzadeh, S. Oliver, S. Rodriguez, and N. Dawood, “On-demand monitoring of construction projects through a game-like hybrid application of BIM and machine learning,” *Automation in Construction*, vol. 110, p. 103012, Feb. 2020, doi: 10.1016/j.autcon.2019.103012.
[28] E. Andenæs, A. Engebø, B. Time, J. Lohne, O. Torp, and T. Kvande, “Perspectives on Quality Risk in the Building Process of Blue-Green Roofs in Norway,” *Buildings*, vol. 10, no. 10, p. 189, 2020, doi: 10.3390/buildings10100189.
[29] V. Saravanan, M. Pourhomayoun, and M. Mazari, “A Proposed Method to Improve Higway Construction Quality Using Machine Learning,” in *2018 International Conference on Computational Science and Computational Intelligence (CSCI)*, Dec. 2018, pp. 1218–1221. doi: 10.1109/CSCI46756.2018.00234.
[30] R. Sacks, M. Girolami, and I. Brilakis, “Building Information Modelling, Artificial Intelligence and Construction Tech,” *Developments in the Built Environment*, vol. 4, p. 100011, Nov. 2020, doi: 10.1016/j.dibe.2020.100011.

[31] M. Al Qady and A. Kandil, “Concept Relation Extraction from Construction Documents Using Natural Language Processing,” *Journal of Construction Engineering and Management*, vol. 136, no. 3, pp. 294–302, Mar. 2010, doi: 10.1061/(ASCE)CO.1943-7862.0000131.
[32] T. Bloch and R. Sacks, “Comparing machine learning and rule-based inferencing for semantic enrichment of BIM models,” *Automation in Construction*, vol. 91, pp. 256–272, Jul. 2018, doi: 10.1016/j.autcon.2018.03.018.
[33] M. Pournader, H. Ghaderi, A. Hassanzadegan, and B. Fahimnia, “Artificial intelligence applications in supply chain management,” *International Journal of Production Economics*, vol. 241, p. 108250, Nov. 2021, doi: 10.1016/j.ijpe.2021.108250.
[34] S. Choudhari and A. Tindwani, “Logistics optimisation in road construction project,” *Construction Innovation*, vol. 17, no. 2, pp. 158–179, Jan. 2017, doi: 10.1108/CI-03-2016-0014.
[35] Y. Fang and S. T. Ng, “Genetic algorithm for determining the construction logistics of precast components,” *Engineering, Construction and Architectural Management*, vol. 26, no. 10, pp. 2289–2306, Jan. 2019, doi: 10.1108/ECAM-09-2018-0386.
[36] A. Darko *et al.*, “Artificial intelligence in the AEC industry : scientometric analysis and visualization of research activities,” *Automation in Construction*, vol. 112, p. 103081, Apr. 2020, doi: 10.1016/j.autcon.2020.103081.
[37] Z. M. Yaseen, Z. H. Ali, S. Q. Salih, and N. Al-Ansari, “Prediction of Risk Delay in Construction Projects Using a Hybrid Artificial Intelligence Model,” *Sustainability*, vol. 12, no. 4, Art. no. 4, Jan. 2020, doi: 10.3390/su12041514.
[38] M. O. Sanni-Anibire, R. M. Zin, and S. O. Olatunji, “Machine learning model for delay risk assessment in tall building projects,” *International Journal of Construction Management*, vol. 22, no. 11, pp. 2134–2143, Aug. 2022, doi: 10.1080/15623599.2020.1768326.
[39] S. Pokharel, T. Roy, and D. Admiraal, “Effects of mass balance, energy balance, and storage-discharge constraints on LSTM for streamflow prediction,” *Environmental Modelling & Software*, vol. 166, p. 105730, Aug. 2023, doi: 10.1016/j.envsoft.2023.105730.
[40] F. Afzal, S. Yunfei, M. Nazir, and S. M. Bhatti, “A review of artificial intelligence based risk assessment methods for capturing complexity-risk interdependencies: Cost overrun in construction projects,” *International Journal of Managing Projects in Business*, vol. 14, no. 2, pp. 300–328, Jan. 2019, doi: 10.1108/IJMPB-02-2019-0047.
[41] S.-S. Lin, S.-L. Shen, A. Zhou, and Y.-S. Xu, “Risk assessment and management of excavation system based on fuzzy set theory and machine learning methods,” *Automation in Construction*, vol. 122, p. 103490, Feb. 2021, doi: 10.1016/j.autcon.2020.103490.
[42] J.-H. Chen, “KNN based knowledge-sharing model for severe change order disputes in construction,” *Automation in Construction*, vol. 17, no. 6, pp. 773–779, Aug. 2008, doi: 10.1016/j.autcon.2008.02.005.
[43] J.-S. Chou and C. Lin, “Predicting Disputes in Public-Private Partnership Projects: Classification and Ensemble Models,” *Journal of Computing in Civil Engineering*, vol. 27, no. 1, pp. 51–60, Jan. 2013, doi: 10.1061/(ASCE)CP.1943-5487.0000197.
[44] W. Lu, J. Lou, C. Webster, F. Xue, Z. Bao, and B. Chi, “Estimating construction waste generation in the Greater Bay Area, China using machine learning,” *Waste Management*, vol. 134, pp. 78–88, Oct. 2021, doi: 10.1016/j.wasman.2021.08.012.
[45] G. Coskuner, M. S. Jassim, M. Zontul, and S. Karateke, “Application of artificial intelligence neural network modeling to predict the generation of domestic, commercial and construction wastes,” *Waste Manag Res*, vol. 39, no. 3, pp. 499–507, Mar. 2021, doi: 10.1177/0734242X20935181.
[46] K. H. Yu, Y. Zhang, D. Li, C. E. Montenegro-Marin, and P. M. Kumar, “Environmental planning based on reduce, reuse, recycle and recover using artificial intelligence,” *Environmental Impact Assessment Review*, vol. 86, p. 106492, Jan. 2021, doi: 10.1016/j.eiar.2020.106492.
[47] M. U. Mehmood, D. Chun, Zeeshan, H. Han, G. Jeon, and K. Chen, “A review of the applications of artificial intelligence and big data to buildings for energy-efficiency and a comfortable indoor living

environment," *Energy and Buildings*, vol. 202, p. 109383, Nov. 2019, doi: 10.1016/j.enbuild.2019.109383.

[48] S. Fathi, R. Srinivasan, A. Fenner, and S. Fathi, "Machine learning applications in urban building energy performance forecasting: A systematic review," *Renewable and Sustainable Energy Reviews*, vol. 133, p. 110287, Nov. 2020, doi: 10.1016/j.rser.2020.110287.

[49] Nasruddin, Sholahudin, P. Satrio, T. M. I. Mahlia, N. Giannetti, and K. Saito, "Optimization of HVAC system energy consumption in a building using artificial neural network and multi-objective genetic algorithm," *Sustainable Energy Technologies and Assessments*, vol. 35, pp. 48–57, Oct. 2019, doi: 10.1016/j.seta.2019.06.002.

[50] C. Debrah, A. P. C. Chan, and A. Darko, "Artificial intelligence in green building," *Automation in Construction*, vol. 137, p. 104192, May 2022, doi: 10.1016/j.autcon.2022.104192.

[51] A. K. Kar, S. K. Choudhary, and V. K. Singh, "How can artificial intelligence impact sustainability: A systematic literature review," *Journal of Cleaner Production*, vol. 376, p. 134120, Nov. 2022, doi: 10.1016/j.jclepro.2022.134120.

[52] J. Seo, H. Park, and S. Choo, "Inference of Drawing Elements and Space Usage on Architectural Drawings Using Semantic Segmentation," *Applied Sciences*, vol. 10, no. 20, Art. no. 20, Jan. 2020, doi: 10.3390/app10207347.

[53] K. Tan, "The Framework of Combining Artificial Intelligence and Construction 3D Printing in Civil Engineering," *MATEC Web Conf.*, vol. 206, p. 01008, 2018, doi: 10.1051/matecconf/201820601008.

[54] T. N. Van and T. N. Quoc, "Research Trends on Machine Learning in Construction Management: A Scientometric Analysis," *Journal of Applied Science and Technology Trends*, vol. 2, no. 03, Art. no. 03, Sep. 2021, doi: 10.38094/jastt203105.

[55] Y. Cao and B. Ashuri, "Predicting the Volatility of Highway Construction Cost Index Using Long Short-Term Memory," *Journal of Management in Engineering*, vol. 36, no. 4, p. 04020020, Jul. 2020, doi: 10.1061/(ASCE)ME.1943-5479.0000784.

[56] N. Semaan and M. Salem, "A deterministic contractor selection decision support system for competitive bidding," *Engineering, Construction and Architectural Management*, vol. 24, no. 1, pp. 61–77, Jan. 2017, doi: 10.1108/ECAM-06-2015-0094.

[57] C. Liu, S. M.E. Sepasgozar, S. Shirowzhan, and G. Mohammadi, "Applications of object detection in modular construction based on a comparative evaluation of deep learning algorithms," *Construction Innovation*, vol. 22, no. 1, pp. 141–159, Jan. 2021, doi: 10.1108/CI-02-2020-0017.

[58] A. Zabin, V. A. González, Y. Zou, and R. Amor, "Applications of machine learning to BIM: A systematic literature review," *Advanced Engineering Informatics*, vol. 51, p. 101474, Jan. 2022, doi: 10.1016/j.aei.2021.101474.

[59] J. Kim, J. Liu, and P. Ghimire, *The Categorization of Virtual Design and Construction Services*. 2019.

[60] S. Mulero-Palencia, S. Álvarez-Díaz, and M. Andrés-Chicote, "Machine Learning for the Improvement of Deep Renovation Building Projects Using As-Built BIM Models," *Sustainability*, vol. 13, no. 12, Art. no. 12, Jan. 2021, doi: 10.3390/su13126576.

[61] S. Paneru, P. Ghimire, A. Kandel, S. Thapa, N. Koirala, and M. Karki, "An Exploratory Investigation of Implementation of Building Information Modeling in Nepalese Architecture–Engineering–Construction Industry," *Buildings*, vol. 13, no. 2, Art. no. 2, Feb. 2023, doi: 10.3390/buildings13020552.

[62] D. Bassir, H. Lodge, H. Chang, J. Majak, and G. Chen, "Application of artificial intelligence and machine learning for BIM: review," *Int. J. Simul. Multidisci. Des. Optim.*, vol. 14, p. 5, 2023, doi: 10.1051/smdo/2023005.

[63] M. Pan, Y. Yang, Z. Zheng, and W. Pan, "Artificial Intelligence and Robotics for Prefabricated and Modular Construction: A Systematic Literature Review," *Journal of Construction Engineering and Management*, vol. 148, no. 9, p. 03122004, Sep. 2022, doi: 10.1061/(ASCE)CO.1943-7862.0002324.

[64] K. You, C. Zhou, and L. Ding, "Deep learning technology for construction machinery and robotics," *Automation in Construction*, vol. 150, p. 104852, Jun. 2023, doi: 10.1016/j.autcon.2023.104852.

[65] T. Bock, “Construction robotics,” *Auton Robot*, vol. 22, no. 3, pp. 201–209, Apr. 2007, doi: 10.1007/s10514-006-9008-5.
[66] H. Oyediran, P. Ghimire, M. Peavy, K. Kim, and P. Barutha, *Robotics Applicability for Routine Operator Tasks in Power Plant Facilities*. 2021. doi: 10.22260/ISARC2021/0091.
[67] B. Meskó and E. J. Topol, “The imperative for regulatory oversight of large language models (or generative AI) in healthcare,” *npj Digit. Med.*, vol. 6, no. 1, Art. no. 1, Jul. 2023, doi: 10.1038/s41746-023-00873-0.
[68] T. Dogru *et al.*, “Generative Artificial Intelligence in the Hospitality and Tourism Industry: Developing a Framework for Future Research,” *Journal of Hospitality & Tourism Research*, p. 10963480231188664, Jul. 2023, doi: 10.1177/10963480231188663.
[69] Y. K. Dwivedi *et al.*, “Opinion Paper: ‘So what if ChatGPT wrote it?’ Multidisciplinary perspectives on opportunities, challenges and implications of generative conversational AI for research, practice and policy,” *International Journal of Information Management*, vol. 71, p. 102642, Aug. 2023, doi: 10.1016/j.ijinfomgt.2023.102642.
[70] F. Fui-Hoon Nah, R. Zheng, J. Cai, K. Siau, and L. Chen, “Generative AI and ChatGPT: Applications, challenges, and AI-human collaboration,” *Journal of Information Technology Case and Application Research*, pp. 1–28, Jul. 2023, doi: 10.1080/15228053.2023.2233814.
[71] A. Kammoun, R. Slama, H. Tabia, T. Ouni, and M. Abid, “Generative Adversarial Networks for Face Generation: A Survey,” *ACM Comput. Surv.*, vol. 55, no. 5, p. 94:1-94:37, Dec. 2022, doi: 10.1145/3527850.
[72] A. N. Wu, R. Stouffs, and F. Biljecki, “Generative Adversarial Networks in the built environment: A comprehensive review of the application of GANs across data types and scales,” *Building and Environment*, vol. 223, p. 109477, Sep. 2022, doi: 10.1016/j.buildenv.2022.109477.
[73] I. Goodfellow *et al.*, “Generative adversarial networks,” *Commun. ACM*, vol. 63, no. 11, pp. 139–144, Oct. 2020, doi: 10.1145/3422622.
[74] A. You, J. K. Kim, I. H. Ryu, and T. K. Yoo, “Application of generative adversarial networks (GAN) for ophthalmology image domains: a survey,” *Eye and Vision*, vol. 9, no. 1, p. 6, Feb. 2022, doi: 10.1186/s40662-022-00277-3.
[75] K. Wang, C. Gou, Y. Duan, Y. Lin, X. Zheng, and F.-Y. Wang, “Generative adversarial networks: introduction and outlook,” *IEEE/CAA Journal of Automatica Sinica*, vol. 4, no. 4, pp. 588–598, 2017, doi: 10.1109/JAS.2017.7510583.
[76] C. Chokwitthaya, E. Collier, Y. Zhu, and S. Mukhopadhyay, “Improving Prediction Accuracy in Building Performance Models Using Generative Adversarial Networks (GANs).” arXiv, Jun. 14, 2019. Accessed: Sep. 05, 2023. [Online]. Available: http://arxiv.org/abs/1906.05767
[77] Y. Yang, K. Zheng, C. Wu, and Y. Yang, “Improving the Classification Effectiveness of Intrusion Detection by Using Improved Conditional Variational AutoEncoder and Deep Neural Network,” *Sensors*, vol. 19, no. 11, Art. no. 11, Jan. 2019, doi: 10.3390/s19112528.
[78] “Autoregressive Models in Deep Learning — A Brief Survey,” ⁂ *George Ho*, Mar. 09, 2019. https://www.georgeho.org/deep-autoregressive-models/ (accessed Sep. 11, 2023).
[79] L. Weng, “What are Diffusion Models?,” Jul. 11, 2021. https://lilianweng.github.io/posts/2021-07-11-diffusion-models/ (accessed Sep. 16, 2023).
[80] L. Weng, “Flow-based Deep Generative Models,” Oct. 13, 2018. https://lilianweng.github.io/posts/2018-10-13-flow-models/ (accessed Sep. 16, 2023).
[81] C. Doersch, “Tutorial on Variational Autoencoders.” arXiv, Jan. 03, 2021. doi: 10.48550/arXiv.1606.05908.
[82] D. P. Kingma and M. Welling, “Auto-Encoding Variational Bayes.” arXiv, Dec. 10, 2022. doi: 10.48550/arXiv.1312.6114.
[83] D. P. Kingma and M. Welling, “An Introduction to Variational Autoencoders,” *MAL*, vol. 12, no. 4, pp. 307–392, Nov. 2019, doi: 10.1561/2200000056.
[84] S. Bond-Taylor, A. Leach, Y. Long, and C. G. Willcocks, “Deep Generative Modelling: A Comparative Review of VAEs, GANs, Normalizing Flows, Energy-Based and Autoregressive

Models," *IEEE Transactions on Pattern Analysis and Machine Intelligence*, vol. 44, no. 11, pp. 7327–7347, Nov. 2022, doi: 10.1109/TPAMI.2021.3116668.
[85] D. Huang *et al.*, "A Variational Autoencoder Based Generative Model of Urban Human Mobility," in *2019 IEEE Conference on Multimedia Information Processing and Retrieval (MIPR)*, Mar. 2019, pp. 425–430. doi: 10.1109/MIPR.2019.00086.
[86] J. M. Davila Delgado and L. Oyedele, "Deep learning with small datasets: using autoencoders to address limited datasets in construction management," *Applied Soft Computing*, vol. 112, p. 107836, Nov. 2021, doi: 10.1016/j.asoc.2021.107836.
[87] V. M. Balmer *et al.*, "Design Space Exploration and Explanation via Conditional Variational Autoencoders in Meta-model-based Conceptual Design of Pedestrian Bridges." arXiv, Nov. 29, 2022. doi: 10.48550/arXiv.2211.16406.
[88] A. Vaswani *et al.*, "Attention is All you Need," in *Advances in Neural Information Processing Systems*, Curran Associates, Inc., 2017. Accessed: Sep. 11, 2023. [Online]. Available: https://proceedings.neurips.cc/paper_files/paper/2017/hash/3f5ee243547dee91fbd053c1c4a845aa-Abstract.html
[89] Y. Bengio, R. Ducharme, and P. Vincent, "A Neural Probabilistic Language Model," in *Advances in Neural Information Processing Systems*, MIT Press, 2000. Accessed: Sep. 11, 2023. [Online]. Available: https://proceedings.neurips.cc/paper_files/paper/2000/hash/728f206c2a01bf572b5940d7d9a8fa4c-Abstract.html
[90] Y. Elfahham, "Estimation and prediction of construction cost index using neural networks, time series, and regression," *Alexandria Engineering Journal*, vol. 58, no. 2, pp. 499–506, Jun. 2019, doi: 10.1016/j.aej.2019.05.002.
[91] J. Ho, A. Jain, and P. Abbeel, "Denoising Diffusion Probabilistic Models," in *Advances in Neural Information Processing Systems*, Curran Associates, Inc., 2020, pp. 6840–6851. Accessed: Sep. 16, 2023. [Online]. Available: https://proceedings.neurips.cc/paper/2020/hash/4c5bcfec8584af0d967f1ab10179ca4b-Abstract.html
[92] A. Kazerouni *et al.*, "Diffusion Models for Medical Image Analysis: A Comprehensive Survey." arXiv, Jun. 03, 2023. doi: 10.48550/arXiv.2211.07804.
[93] F.-A. Croitoru, V. Hondru, R. T. Ionescu, and M. Shah, "Diffusion Models in Vision: A Survey," *IEEE Transactions on Pattern Analysis and Machine Intelligence*, vol. 45, no. 9, pp. 10850–10869, Sep. 2023, doi: 10.1109/TPAMI.2023.3261988.
[94] L. Dinh, D. Krueger, and Y. Bengio, "NICE: Non-linear Independent Components Estimation," *arXiv.org*, Oct. 30, 2014. https://arxiv.org/abs/1410.8516v6 (accessed Sep. 16, 2023).
[95] M. Kumar *et al.*, *VideoFlow: A Flow-Based Generative Model for Video*. 2019.
[96] L. Dinh, J. Sohl-Dickstein, and S. Bengio, "Density estimation using Real NVP," *arXiv.org*, May 27, 2016. https://arxiv.org/abs/1605.08803v3 (accessed Sep. 16, 2023).
[97] J. Zheng and M. Fischer, "BIM-GPT: a Prompt-Based Virtual Assistant Framework for BIM Information Retrieval".
[98] S. A. Prieto, E. T. Mengiste, and B. García de Soto, "Investigating the Use of ChatGPT for the Scheduling of Construction Projects," *Buildings*, vol. 13, no. 4, Art. no. 4, Apr. 2023, doi: 10.3390/buildings13040857.
[99] H. A. Mohamed Hassan, E. Marengo, and W. Nutt, "A BERT-Based Model for Question Answering on Construction Incident Reports," in *Natural Language Processing and Information Systems*, P. Rosso, V. Basile, R. Martínez, E. Métais, and F. Meziane, Eds., in Lecture Notes in Computer Science. Cham: Springer International Publishing, 2022, pp. 215–223. doi: 10.1007/978-3-031-08473-7_20.
[100] S. Moon, S. Chi, and S.-B. Im, "Automated detection of contractual risk clauses from construction specifications using bidirectional encoder representations from transformers (BERT)," *Automation in Construction*, vol. 142, p. 104465, Oct. 2022, doi: 10.1016/j.autcon.2022.104465.

[101] S. Chung, S. Moon, J. Kim, J. Kim, S. Lim, and S. Chi, “Comparing natural language processing (NLP) applications in construction and computer science using preferred reporting items for systematic reviews (PRISMA),” *Automation in Construction*, vol. 154, p. 105020, Oct. 2023, doi: 10.1016/j.autcon.2023.105020.

[102] H. You, Y. Ye, T. Zhou, Q. Zhu, and J. Du, “Robot-Enabled Construction Assembly with Automated Sequence Planning based on ChatGPT: RoboGPT.” arXiv, Apr. 21, 2023. doi: 10.48550/arXiv.2304.11018.

[103] “AI Caucus Leaders Introduce Bipartisan Bill to Expand Access to AI Research,” *Congresswoman Anna Eshoo*, Jul. 28, 2023. http://eshoo.house.gov/media/press-releases/ai-caucus-leaders-introduce-bipartisan-bill-expand-access-ai-research (accessed Aug. 16, 2023).

[104] L. Floridi and M. Chiriatti, “GPT-3: Its Nature, Scope, Limits, and Consequences,” *Minds & Machines*, vol. 30, no. 4, pp. 681–694, Dec. 2020, doi: 10.1007/s11023-020-09548-1.

[105] “GPT-3,” *Wikipedia*. Aug. 13, 2023. Accessed: Aug. 26, 2023. [Online]. Available: https://en.wikipedia.org/w/index.php?title=GPT-3&oldid=1170092033

[106] “GPT-4.” https://openai.com/gpt-4 (accessed Aug. 26, 2023).

[107] A. Chowdhery *et al.*, “PaLM: Scaling Language Modeling with Pathways.” arXiv, Oct. 05, 2022. Accessed: Aug. 26, 2023. [Online]. Available: http://arxiv.org/abs/2204.02311

[108] “Google AI PaLM 2 – Google AI.” https://ai.google/discover/palm2/ (accessed Aug. 26, 2023).

[109] H. Touvron *et al.*, “LLaMA: Open and Efficient Foundation Language Models.” arXiv, Feb. 27, 2023. Accessed: Aug. 26, 2023. [Online]. Available: http://arxiv.org/abs/2302.13971

[110] S. Zhang *et al.*, “OPT: Open Pre-trained Transformer Language Models.” arXiv, Jun. 21, 2022. Accessed: Aug. 26, 2023. [Online]. Available: http://arxiv.org/abs/2205.01068

[111] A. Sha, “12 Best Large Language Models (LLMs) in 2023,” *Beebom*, Jun. 27, 2023. https://beebom.com/best-large-language-models-llms/ (accessed Aug. 26, 2023).

[112] KORKRID (KYLE), “Data Behind the Large Language Models (LLM), GPT, and Beyond,” *Medium*, Apr. 11, 2023. https://kyleake.medium.com/data-behind-the-large-language-models-llm-gpt-and-beyond-8b34f508b5de (accessed Aug. 26, 2023).

[113] F. Heimerl, S. Lohmann, S. Lange, and T. Ertl, “Word Cloud Explorer: Text Analytics Based on Word Clouds,” in *2014 47th Hawaii International Conference on System Sciences*, Jan. 2014, pp. 1833–1842. doi: 10.1109/HICSS.2014.231.

[114] A. I. Kabir, K. AHMED, and R. Karim, “Word Cloud and Sentiment Analysis of Amazon Earphones Reviews with R Programming Language,” *Informatica Economica*, vol. 24, pp. 55–71, Dec. 2020, doi: 10.24818/issn14531305/24.4.2020.05.

[115] L. V. T. Vencer, H. Bansa, and A. R. Caballero, “Data and Sentiment Analysis of Monkeypox Tweets using Natural Language Toolkit (NLTK),” in *2023 8th International Conference on Business and Industrial Research (ICBIR)*, May 2023, pp. 392–396. doi: 10.1109/ICBIR57571.2023.10147684.

[116] J. Thanaki, *Python Natural Language Processing*. Packt Publishing Ltd, 2017.

[117] M. Birjali, M. Kasri, and A. Beni-Hssane, “A comprehensive survey on sentiment analysis: Approaches, challenges and trends,” *Knowledge-Based Systems*, vol. 226, p. 107134, Aug. 2021, doi: 10.1016/j.knosys.2021.107134.

[118] G. Xu, Y. Meng, X. Qiu, Z. Yu, and X. Wu, “Sentiment Analysis of Comment Texts Based on BiLSTM,” *IEEE Access*, vol. 7, pp. 51522–51532, 2019, doi: 10.1109/ACCESS.2019.2909919.

[119] W. Medhat, A. Hassan, and H. Korashy, “Sentiment analysis algorithms and applications: A survey,” *Ain Shams Engineering Journal*, vol. 5, no. 4, pp. 1093–1113, Dec. 2014, doi: 10.1016/j.asej.2014.04.011.

[120] X. Xu, L. Ma, and L. Ding, “A Framework for BIM-Enabled Life-Cycle Information Management of Construction Project,” *International Journal of Advanced Robotic Systems*, vol. 11, no. 8, p. 126, Aug. 2014, doi: 10.5772/58445.

[121] W. Hu, "Information Lifecycle Modeling Framework for Construction Project Lifecycle Management," in *2008 International Seminar on Future Information Technology and Management Engineering*, Nov. 2008, pp. 372–375. doi: 10.1109/FITME.2008.142.
[122] B. Succar, "Building information modelling framework: A research and delivery foundation for industry stakeholders," *Automation in Construction*, vol. 18, no. 3, pp. 357–375, May 2009, doi: 10.1016/j.autcon.2008.10.003.
[123] B. Becerik-Gerber, F. Jazizadeh, N. Li, and G. Calis, "Application Areas and Data Requirements for BIM-Enabled Facilities Management," *Journal of Construction Engineering and Management*, vol. 138, no. 3, pp. 431–442, Mar. 2012, doi: 10.1061/(ASCE)CO.1943-7862.0000433.
[124] A. Saka *et al.*, "GPT Models in Construction Industry: Opportunities, Limitations, and a Use Case Validation." arXiv, May 30, 2023. doi: 10.48550/arXiv.2305.18997.
[125] T. Brown *et al.*, "Language Models are Few-Shot Learners," in *Advances in Neural Information Processing Systems*, Curran Associates, Inc., 2020, pp. 1877–1901. Accessed: Aug. 26, 2023. [Online]. Available: https://proceedings.neurips.cc/paper_files/paper/2020/hash/1457c0d6bfcb4967418bfb8ac142f64a-Abstract.html
[126] S. Amershi *et al.*, "Guidelines for Human-AI Interaction," in *Proceedings of the 2019 CHI Conference on Human Factors in Computing Systems*, in CHI '19. New York, NY, USA: Association for Computing Machinery, May 2019, pp. 1–13. doi: 10.1145/3290605.3300233.
[127] R. Goel, S. Vashisht, A. Dhanda, and S. Susan, "An Empathetic Conversational Agent with Attentional Mechanism," in *2021 International Conference on Computer Communication and Informatics (ICCCI)*, Jan. 2021, pp. 1–4. doi: 10.1109/ICCCI50826.2021.9402337.
[128] C.-H. Kuo, C.-T. Chen, S.-J. Lin, and S.-H. Huang, "Improving Generalization in Reinforcement Learning–Based Trading by Using a Generative Adversarial Market Model," *IEEE Access*, vol. 9, pp. 50738–50754, 2021, doi: 10.1109/ACCESS.2021.3068269.
[129] Y. Li, Q. Pan, S. Wang, T. Yang, and E. Cambria, "A Generative Model for category text generation," *Information Sciences*, vol. 450, pp. 301–315, Jun. 2018, doi: 10.1016/j.ins.2018.03.050.
[130] J. E. Zini and M. Awad, "On the Explainability of Natural Language Processing Deep Models," *ACM Comput. Surv.*, vol. 55, no. 5, p. 103:1-103:31, Dec. 2022, doi: 10.1145/3529755.
[131] R. Bommasani *et al.*, "On the Opportunities and Risks of Foundation Models." arXiv, Jul. 12, 2022. doi: 10.48550/arXiv.2108.07258.
[132] A. B. Saka, L. O. Oyedele, L. A. Akanbi, S. A. Ganiyu, D. W. M. Chan, and S. A. Bello, "Conversational artificial intelligence in the AEC industry: A review of present status, challenges and opportunities," *Advanced Engineering Informatics*, vol. 55, p. 101869, Jan. 2023, doi: 10.1016/j.aei.2022.101869.
[133] D. U. Patton, A. Y. Landau, and S. Mathiyazhagan, "ChatGPT for Social Work Science: Ethical Challenges and Opportunities," *Journal of the Society for Social Work and Research*, pp. 000–000, Aug. 2023, doi: 10.1086/726042.
[134] A. Piñeiro-Martín, C. García-Mateo, L. Docío-Fernández, and M. del C. López-Pérez, "Ethical Challenges in the Development of Virtual Assistants Powered by Large Language Models," *Electronics*, vol. 12, no. 14, Art. no. 14, Jan. 2023, doi: 10.3390/electronics12143170.
[135] P. Liu, W. Yuan, J. Fu, Z. Jiang, H. Hayashi, and G. Neubig, "Pre-train, Prompt, and Predict: A Systematic Survey of Prompting Methods in Natural Language Processing," *ACM Comput. Surv.*, vol. 55, no. 9, p. 195:1-195:35, Jan. 2023, doi: 10.1145/3560815.
[136] B. Meskó and E. J. Topol, "The imperative for regulatory oversight of large language models (or generative AI) in healthcare," *npj Digit. Med.*, vol. 6, no. 1, Art. no. 1, Jul. 2023, doi: 10.1038/s41746-023-00873-0.
[137] "(1) Activity | Dr. Bradley Hyatt | LinkedIn." https://www.linkedin.com/in/bradleyhyatt/recent-activity/all/ (accessed Sep. 18, 2023).